\documentclass[runningheads]{llncs}

\usepackage{eccv}

\usepackage{eccvabbrv}

\usepackage{graphicx}
\usepackage{booktabs}
\usepackage{color}
\usepackage{colortbl}
\usepackage{tcolorbox}
\usepackage{mdframed}
\usepackage[hidelinks]{hyperref}

\usepackage[accsupp]{axessibility}  %

\definecolor{amber}{rgb}{1.0, 0.49, 0.0}
\definecolor{amethyst}{rgb}{0.6, 0.4, 0.8}
\definecolor{asparagus}{rgb}{0.53, 0.66, 0.42}
\definecolor{floralwhite}{rgb}{1.0, 0.98, 0.94}
\definecolor{ghostwhite}{rgb}{0.97, 0.97, 1.0}
\definecolor{americanrose}{rgb}{1.0, 0.01, 0.24}
\definecolor{aqua}{rgb}{0.0, 1.0, 1.0}
\definecolor{electricpurple}{rgb}{0.75, 0.0, 1.0}
\definecolor{electricultramarine}{rgb}{0.25, 0.0, 1.0}
\definecolor{palatinateblue}{rgb}{0.15, 0.23, 0.89}
\definecolor{persiangreen}{rgb}{0.0, 0.65, 0.58}
\definecolor{turquoiseblue}{rgb}{0.0, 1.0, 0.94}
\definecolor{turquoise}{rgb}{0.19, 0.84, 0.78}
\definecolor{blue-green}{rgb}{0.0, 0.87, 0.87}
\definecolor{deepcarminepink}{rgb}{0.94, 0.19, 0.22}
\definecolor{vividcerise}{rgb}{0.85, 0.11, 0.51}
\definecolor{rose}{rgb}{1.0, 0.0, 0.5}
\definecolor{zaffre}{rgb}{0.0, 0.08, 0.66}
\definecolor{warmblack}{rgb}{0.0, 0.26, 0.26}
\definecolor{winter0.0}{rgb}{0.0, 0.0, 1.0}
\definecolor{winter0.2}{rgb}{0.0, 0.2, 0.9}
\definecolor{winter0.4}{rgb}{0.0, 0.4, 0.8}
\definecolor{winter0.6}{rgb}{0.0, 0.6, 0.7}
\definecolor{winter0.8}{rgb}{0.0, 0.8, 0.6}
\definecolor{winter1.0}{rgb}{0.0, 1.0, 0.5}

\definecolor{mayablue}{rgb}{0.45, 0.76, 0.98}

\definecolor{highlighteq}{rgb}{0.44, 0.16, 0.39}

\usepackage{amsmath}
\usepackage{amssymb}
\usepackage{algorithm}
\usepackage{algorithmic}
\usepackage{mathtools}
\usepackage{adjustbox}
\usepackage{hyperref}
\usepackage{multirow}
\usepackage{wrapfig}
\usepackage{tikz}
\usepackage{microtype}
\usepackage{graphicx}
\usepackage{subcaption}
\usepackage{booktabs} %
\usepackage{caption} 
\usepackage{mathrsfs} 
\usepackage{pifont}
\usepackage{xcolor}

\usepackage{orcidlink}
\newcommand{\framework}{\texttt{{\textcolor{blue}{AVReCAP}}}}

\begin{document}

\title{Audio-Visual Continual Test-Time Adaptation \textit{without} Forgetting} 

\titlerunning{\texttt{{AVReCAP}}}

\author{Sarthak Kumar Maharana\inst{1}\thanks{Work primarily done during an internship at Dolby Laboratories.} \and Akshay Mehra\inst{2} \and
Bhavya Ramakrishna\inst{2} \and Yunhui Guo\inst{1} \and Guan-Ming Su\inst{2}}

\authorrunning{S.K Maharana et al.}

\institute{The University of Texas at Dallas, Richardson, TX 75080, USA \and
Dolby Laboratories, Inc., San Francisco, CA 94103, USA\\
\email{sarthak.maharana@utdallas.edu, akshay.mehra@dolby.com}\\
Project page: \url{https://sarthaxxxxx.github.io/AVReCAP/index.html}}

\maketitle

\begin{abstract}
Audio-visual continual test-time adaptation involves continually adapting a source audio-visual model at test-time, to unlabeled non-stationary domains, where either or both modalities can be distributionally shifted, which hampers online cross-modal learning and eventually leads to poor accuracy. While previous works have tackled this problem, we find that SOTA methods suffer from \textit{catastrophic forgetting} where the model's performance drops well below even the source model due to continual parameter updates at test-time. In this work, we first show that adapting only the modality fusion layer to a target domain not only improves performance on that domain but can also enhance performance on subsequent domains. Based on this strong cross-task transferability of the fusion layer's parameters, we propose a method, $\framework$, that improves test-time performance of the models without access to any source data. Our approach works by using a selective parameter retrieval mechanism that dynamically retrieves the best fusion layer parameters from a buffer using only a small batch of test data. These parameters are then integrated into the model, adapted to the current test distribution, and saved back for future use. Extensive experiments on benchmark datasets involving unimodal and bimodal corruptions show our proposed $\framework$ significantly outperforms existing methods while minimizing catastrophic forgetting.  
\end{abstract}

\section{Introduction}
\label{intro}

The development of audio-visual models has accelerated rapidly in recent years \cite{likhosherstov2021polyvit, akbari2021vatt, gong2022uavm, gong2022contrastive, huang2023mavil, kim2024equiav, araujo2025cav}. However, most prior work evaluates these models under in-distribution settings that closely mirror their pre-training data. In real-world deployment, models operate under evolving test-time distribution shifts that significantly hinder generalization \cite{taori2020measuring, liu2022empirical}. Consider safety-critical scenarios such as self-driving cars equipped with audio-visual sensors. As the vehicle encounters changing visual conditions (\eg \textit{snow}, \textit{fog}) or diverse acoustic environments, cross-modal associations learned during training may no longer hold, leading to degraded perception. The challenge becomes even more severe when both modalities shift simultaneously, such as in a robot deployed for scene understanding in a \textit{crowded} street \cite{maharana2025texttt}.

\begin{figure}
    \centering
    \includegraphics[trim={55 200 55 160},clip,width=0.6\linewidth]{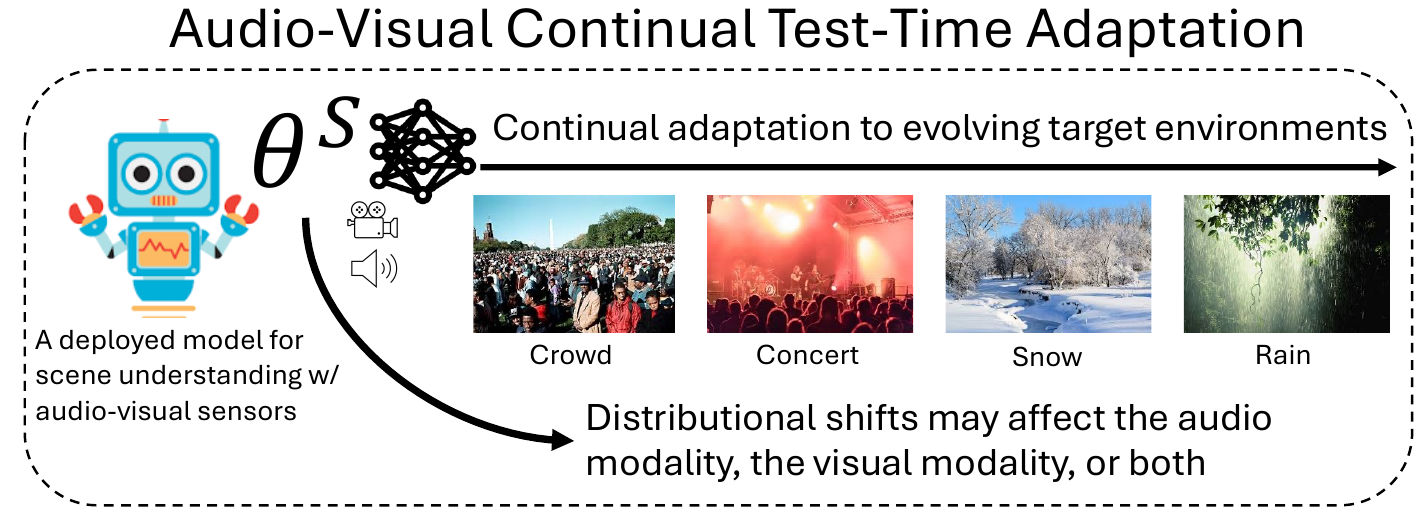}
    \caption{We illustrate \textbf{source-free audio-visual continual test-time adaptation} using an example of a deployed model (say, in a robot) with audio-visual sensors for scene understanding. Starting from a source model parameterized by $\theta^S$, it encounters a sequence of evolving target environments where distributional shifts may affect the audio modality, the visual modality, or both, motivating continual adaptation at test-time. The goal is to maintain robust performance at test-time without access to the source data and task boundaries.} %
    \vspace{-15pt}
    \label{fig:demo}
\end{figure}

Test-time adaptation (TTA) has emerged as an effective paradigm for improving model robustness under distribution shifts \cite{wang2020tent, liang2025comprehensive}. In realistic scenarios with privacy and real-time constraints \cite{mai2022online}, source data is unavailable, and adaptation must be efficient. Consequently, updates are typically limited to a \textit{single forward pass} per test sample \cite{niu2022efficient, danilowski2025botta}, while ensuring that source knowledge is not catastrophically forgotten \cite{goodfellow2013empirical}. Although there is a long TTA literature on vision models \cite{niu2022efficient, niu2023towards, chen2022contrastive, choi2022improving, zhang2022memo}, it has been recently gaining traction for audio-visual models too \cite{yang2024test, guo2025smoothing, li2025bridging, wangpartition, maharana2025texttt}. We name such methods as AV-TTA.

\begin{figure}[t!]
    \centering
    \includegraphics[trim={10 10 5 5},clip,width=0.6\linewidth]{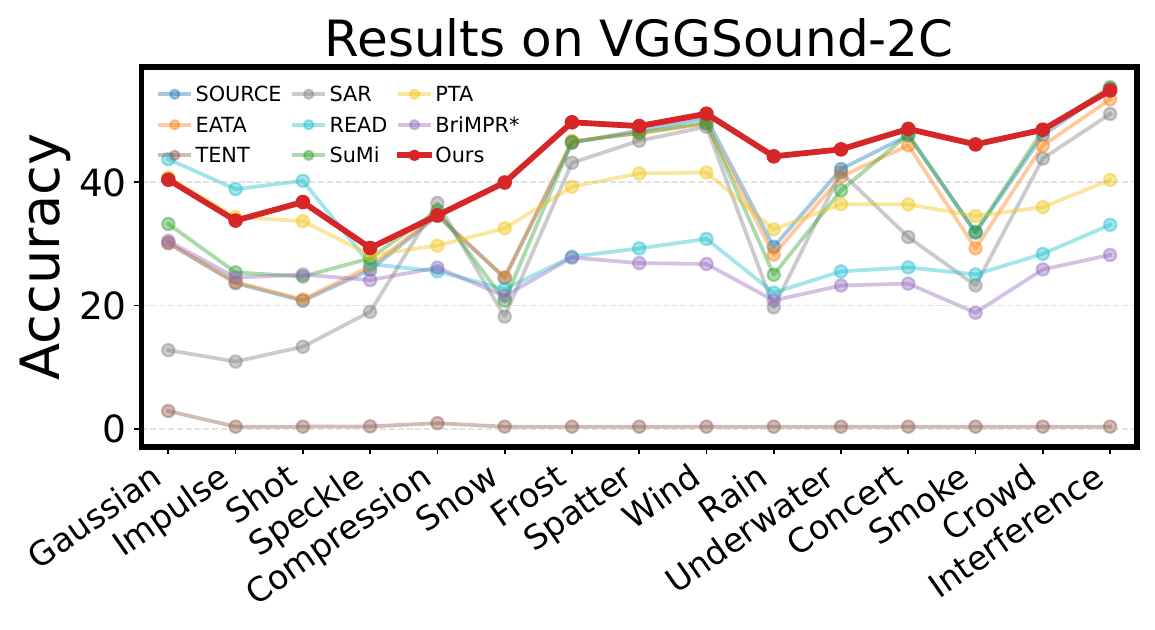}
    \caption{\textbf{\textcolor{Red}{\texttt{AVReCAP}} achieves SOTA performance on audio-visual CTTA.} We report task-wise accuracy on VGGSound-2C \cite{maharana2025texttt} at a severity level of 5 under correlated bimodal corruptions in the continual setting.  CAV-MAE \cite{gong2022contrastive} is used as the \textcolor{MidnightBlue}{\textbf{SOURCE}} model. We extend TTA (\textcolor{RawSienna}{\textbf{TENT}}, \textcolor{orange}{\textbf{EATA}}, \textcolor{Gray}{\textbf{SAR}}) and AV-TTA (\textcolor{SkyBlue}{\textbf{READ}}, \textcolor{yellow}{\textbf{PTA}}, \textcolor{YellowGreen}{\textbf{SuMi}}, \textcolor{Purple}{\textbf{BriMPR*}}) methods to the continual setting. Existing AV-TTA methods struggle under severe correlated bimodal corruptions.}
    \label{fig:error}
    \vspace{-20pt}
\end{figure}

But, this single-domain assumption is unrealistic in real-world deployment, where models face dynamic and unpredictable domain changes without explicit task boundaries \cite{wang2022continual}. This setting, termed continual TTA or CTTA, introduces two key challenges: \textbf{\textit{catastrophic forgetting}} of the source knowledge due to continual parameter updates across domains, and \textbf{\textit{error accumulation}}, where miscalibrated \cite{guo2017calibration} pseudo-labels compound over time. Although CTTA has been widely studied in vision-only models \cite{song2023ecotta, wang2024continual, maharana2025palm, zhang2024dpcore}, it is substantially harder in such multimodal settings. Shifts may affect audio and visual modalities differently, creating a modality gap \cite{maharana2025texttt} that hinders cross-modal alignment \cite{dong2023simmmdg}. Continual updates under such bimodal shifts further amplify error accumulation, leading to progressive performance degradation (see Figure \ref{fig:error}). We use the terms distribution, domain, and corruption interchangeably.

In this work, we study \textit{source-free} audio-visual continual test-time adaptation under both unimodal and challenging bimodal corruptions, as illustrated in Figure \ref{fig:demo}. Although existing AV-TTA methods can be extended to the continual setting, they exhibit limitations. Prior work \cite{maharana2025texttt} shows that, even in single-domain AV-TTA, READ \cite{yang2024test} suffers from increasing attention imbalance between modality tokens under strong bimodal corruptions, causing performance degradation. We observe a consequence of this in the CTTA setting: when extending READ to VGGSound-2C \cite{maharana2025texttt}, performance initially improves but progressively declines as new domains arrive (see Figure \ref{fig:error}). Furthermore, BriMPR \cite{li2025bridging}, an AV-TTA method which relies on source data, degrades significantly when source access is removed in our continual setting; we denote this source-free variant as BriMPR*.

With the dual goals of minimizing source knowledge forgetting and enhancing test-time performance, we propose a new framework \textbf{$\framework$}. In \S \ref{analysis}, we present a key observation: adapting only the attention weights of the fusion layer to a single target domain not only improves performance over the source model on related domains within the same category, but also achieves better or competitive improvements on unseen domains of other categories. This reveals both intra- and cross-task transferability of the adapted fusion parameters. For the continual setting, however, this suggests that, at any time-step, we can retrieve \textit{similar} parameters from past time-steps and reuse them for further adaptation, instead of the continual overwriting of shared parameters that could result in poor accuracy (see Figure \ref{fig:error}). In \S \ref{method:retrieval}, we leverage this core insight and propose a selective parameter retrieval scheme to obtain the most relevant parameter state for adaptation. We introduce a shared buffer to store ``snapshots" of the attention weights, inspired by replay-based methods \cite{rebuffi2017icarl, chaudhry2018efficient, chaudhry2019tiny}, but adhering to strict data privacy rules \cite{shokri2015privacy} by storing only modality-specific input-level statistics and the attention weights, without retaining or revisiting past domain data. At every time-step, we compute current modality-specific statistics to guide the selection. To bound buffer growth (\S \ref{method:buffer_}), we merge statistically similar buffer elements. This effectively avoids overfitting and largely minimizes \textit{catastrophic forgetting} (see Figure \ref{fig:cf}).

Our \textbf{contributions} are as follows: 1) We comprehensively study source-free audio-visual CTTA under unimodal and bimodal corruptions. 2) We show that the attention fusion layer shows intra- and cross-task transferability, suggesting that parameters from previous time-steps can be selectively retrieved and used for adaptation (see \S \ref{analysis}). Our method $\framework$ then proposes maintaining a buffer, under a memory budget, to selectively retrieve such parameters based on a proposed criterion. Crucially, $\framework$ substantially mitigates \textbf{catastrophic forgetting}, incurring only a \textbf{2.9\%} performance drop compared to \textbf{27.9\%} with READ on a difficult test set like VGGSound-2C, involving 15 bimodal corruptions. 3) Our method outperforms all baselines for audio-visual CTTA across datasets.

\section{Related Work}
\label{related}

\noindent \textbf{Audio-Visual Test-Time Adaptation (AV-TTA).} TTA aims to adapt a source model to unlabeled target data without access to source samples, mitigating source–target domain gaps under privacy and real-time constraints \cite{wang2020tent, chen2022contrastive, niu2022efficient, niu2023towards, niu2024test}. Typically, adaptation is performed independently per domain with a single forward pass per sample \cite{mai2022online}.
TENT \cite{wang2020tent} minimizes the Shannon entropy of model predictions by fine-tuning the affine parameters of normalization layers. EATA \cite{niu2022efficient} penalizes high entropy samples while minimizing the entropy. SAR \cite{niu2023towards} identified that model adaptation is affected by samples with large gradients.

TTA has also been widely studied in audio-visual settings \ie AV-TTA. The first work, READ \cite{yang2024test}, proposed adapting the attention weights of the joint encoder to ensure robust cross-modal fusion. SuMi \cite{guo2025smoothing} employs a mutual information sharing loss to enhance the alignment between modalities. PTA \cite{wangpartition} argues that under bimodal domain shifts, the source model results in overlapping class representations, leading to biased predictions. PTA partitions samples by prediction bias and jointly reweighs bias and confidence. BriMPR \cite{li2025bridging} introduces modality-specific prompts to enhance re-alignment. While prompts are optimized for a specific domain, a layer-wise discrepancy loss is minimized between the statistics of the intermediate features and their corresponding source features.

\noindent \textbf{Continual TTA (CTTA).} The research community has extensively studied CTTA in the visual domain \cite{maharana2026continual, wang2022continual, gong2022note, song2023ecotta, gan2023decorate, liu2023vida, wang2024continual, maharana2025palm, zhang2024dpcore}. However, CTTA for audio-visual data has been relatively underexplored. We notice that only PTA \cite{wangpartition} and BriMPR \cite{li2025bridging} provide limited extensions of their respective AV-TTA methods to the continual setting (see their Appendices). \cite{zhang2025analytic} also studies audio-visual CTTA assuming complete access to the source data. A comprehensive study of source-free audio–visual CTTA is still missing, which is the primary focus of this paper.

\vspace{5pt}

\noindent \textit{\underline{Remarks.}}\label{remark} While promising in performance, prior vision-based CTTA methods \cite{song2023ecotta, liu2023vida, zhang2024dpcore} use a small amount of source data to warm-start meta-networks/adapters or guide visual prompt tuning \cite{jia2022visual} at test-time, respectively. Though it might be a common practice, under \textit{strict} real-time and privacy-constrained settings, even limited source access may be unavailable. Similarly, BriMPR reports using only 32 source samples. Interestingly, we notice that when the unlabeled source data is absent, the performance drastically drops, as in Figure \ref{fig:error}. Moving forward, we abide by strict test-time settings and provide a realistic source-free audio-visual CTTA method, and so, do not compare against \cite{zhang2025analytic}.

\section{Preliminaries}
\label{prelim}

\noindent \textbf{Notations.} 
Let $f_a$ and $f_v$ denote two unimodal transformer encoders that process audio and visual inputs, respectively. These are followed by a joint transformer encoder $f_j$, which operates on the concatenated audio-visual output tokens, and a final prediction head $h$ for classification. The total parameter set is $\theta^S$ = \{$\theta_a$, $\theta_v$, $\theta_j$, $\theta_h$\}, where $\theta_a$, $\theta_v$, $\theta_j$, and $\theta_h$ are the parameters of the mentioned modules, respectively. In essence, let $z_a = f_a(x_a; \theta_a)$ where $z_a \in \mathbb{R}^{T_a \times D}$ be the audio tokens and $z_v = f_v(x_v; \theta_v)$ where $z_v \in \mathbb{R}^{T_v \times D}$ be the visual tokens. $T_a$ and $T_v$ represent the respective number of output tokens and $D$ refers to the embedding dimension (usually $D=768$). For cross-modal fusion, these features are concatenated along the token dimension to form $z_c = [z_a; z_v]$, resulting in a joint representation $z_c \in \mathbb{R}^{(T_a + T_v) \times D}$. The final logits, mapped to the concept space, are \textit{l} = $h(f_j(z_c;\theta_j);\theta_h)$. $x_a$ may represent raw audio or its corresponding spectrogram, while $x_v$ denotes the corresponding video frames. Specific to CAV-MAE \cite{gong2022contrastive}, there are 11 attention blocks in each of $f_a$ and $f_v$ and 1 attention block in $f_j$. We denote the weights of the query, key, and value projection matrices in $f_j$ as $\mathcal{W_Q}$, $\mathcal{W_K}$, and $\mathcal{W_V}$ $\in$ $\mathbb{R}^{D \times D}$, respectively.

\noindent \textbf{Problem Setting.}\label{setting} We begin with a model pre-trained on a source domain $\mathcal{T}^S = \{(x_{a,i}^S, x_{v,i}^S, y_i^S)\}_{i=1}^n$, where $x_i^S = (x_{a,i}^S, x_{v,i}^S)$ represents a pair of source audio and visual inputs and $y_{i}^S$ denotes the corresponding ground-truth labels. In the CTTA setting \cite{wang2022continual}, this source model encounters a sequence of $U$ target tasks/domains at test-time, $\mathcal{T} = \{\mathcal{T}_u\}_{u=1}^U$. The objective is to continually adapt the model to each incoming domain in an online fashion, without any parameter reset. For a given task $\mathcal{T}_u$, the model receives a stream of $B$ unlabeled batches. At each time step $t$, the input is a multimodal batch $x^t = \{x_a^t, x_v^t\}$, with labels unknown. We assume a strictly online setting where each batch is available only once for a \textit{single forward pass}. Crucially, the distribution of the source domain differs from the target domain, \ie $p(x_i^S)$ $\neq$ $p(x_i^t)$ but $p(y_i^S)$ $=$ $p(y_i^t)$ for an \textit{i$^{th}$} sample. Target domains also differ from one another. Throughout, the task boundaries are unknown. We assume that the sequence of tasks at test-time are disjoint, \ie $\mathcal{T}_u \cap \left( \bigcup_{i=1}^{u-1} \mathcal{T}_i \right)=\emptyset$. Each $\mathcal{T}_u$ introduces a unique domain that has not been seen earlier, but the concept or label spaces remain the same. To be noted, \textit{either the audio, visual, or both modalities can be shifted/corrupted, \ie unimodal or bimodal corruptions, respectively.}

\section{Method}
\label{proposed}

\begin{figure*}[ht!]
    \centering
    \begin{subfigure}[t]{0.5\textwidth}
        \centering
        \includegraphics[width=\linewidth]{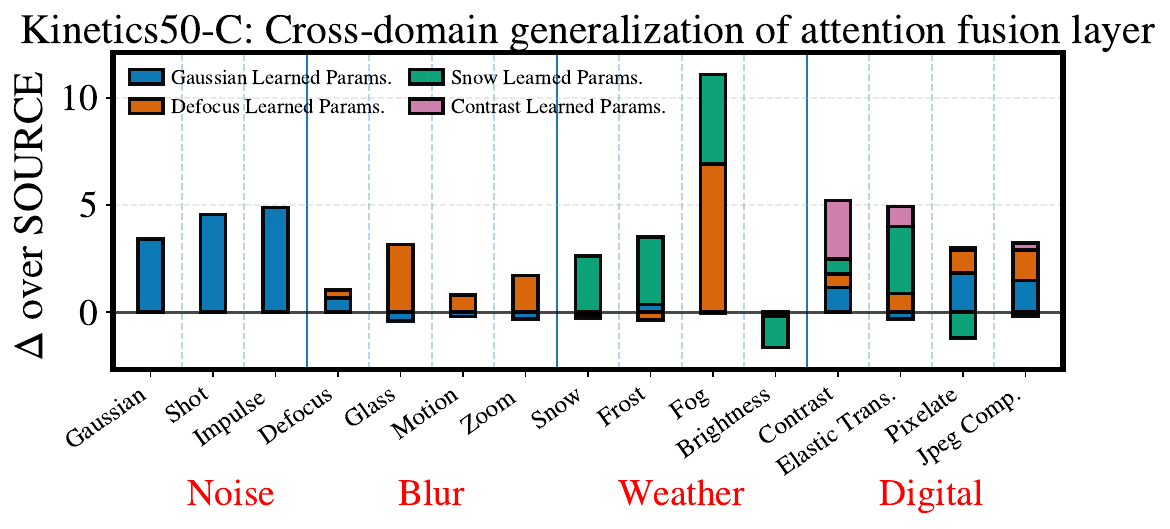}
        \caption{Kinetics50-C with visual corruptions.}
        \label{fig:attn_ks}
    \end{subfigure}\hfill
    \begin{subfigure}[t]{0.5\textwidth}
        \centering
        \includegraphics[width=\linewidth]{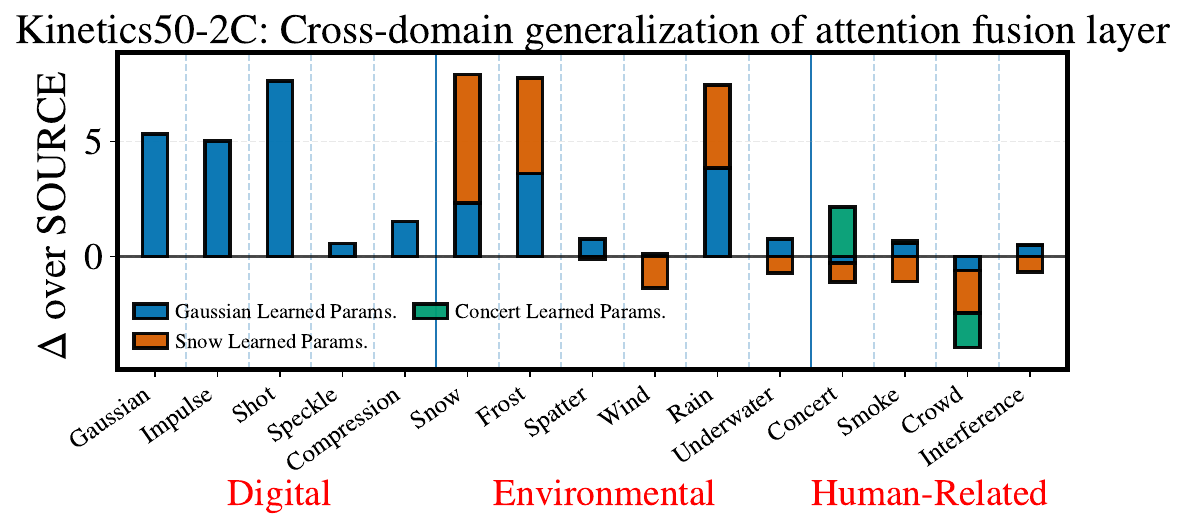}
        \caption{Kinetics50-2C with bimodal corruptions.}
        \label{fig:attn_ks2c}
    \end{subfigure}

    \caption{
     \textbf{Attention fusion layer adapted on a single domain successfully transfers, achieving performance exceeding or comparable to the source model, motivating us to store parameter snapshots in a buffer that can be reused during audio-visual CTTA.} We adapt the projection matrices $\{\mathcal{W}_Q, \mathcal{W}_K, \mathcal{W}_V\}$ of the joint encoder $f_j$ of a pre-trained CAV-MAE (SOURCE) \cite{gong2022contrastive} on the first unseen domain of each corruption category, following READ \cite{yang2024test}. This adapted state is then frozen and inferred on the remaining sequence of unseen domains. We report the accuracy change $\Delta$ (in \%) over SOURCE.}
    \label{fig:attn_generalization}
    \vspace{-18pt}
\end{figure*}

In this section, we introduce our proposed method $\framework$ for audio-visual continual test-time adaptation. Before diving in, we provide a motivating rationale. 

\subsection{Motivation}
\label{analysis}
Cross-domain generalization \cite{huang2020self, wang2019learning, zhang2022towards, li2017deeper} is a well-studied problem in computer vision that aims to improve performance on unseen and unlabeled test domains by training models on diverse labeled source domains. Although most previous works focus on improving generalization through training \cite{carlucci2019domain, zhou2020learning}, we slightly abuse the setting here as the source data is absent. In this section, we present an interesting observation by studying cross-domain generalization at test-time. \textit{Specifically, we show that when adapting the attention fusion layer, \ie $(\mathcal{W}_Q, \mathcal{W}_K, \mathcal{W}_V)$ of a source model on an unlabeled test/target domain of a corruption category, the learned model parameters exhibit improved/competitive transfer to other unseen test/target domains within the same category.} We illustrate the results in Figure \ref{fig:attn_generalization}.

We evaluate on Kinetics50-C \cite{yang2024test} with visual-only corruptions \cite{hendrycks2019benchmarking}, grouped into \textit{Noise}, \textit{Blur}, \textit{Weather}, and \textit{Digital} categories \cite{hendrycks2019benchmarking}. Starting from a CAV-MAE source model \cite{gong2022contrastive}, pre-trained on Kinetics50, we adapt only the attention fusion layer ($\mathcal{W_Q}$, $\mathcal{W_K}$, and $\mathcal{W_V}$ of $f_j$) following READ \cite{yang2024test}, with pseudo-labels based on the model's predictions. For each category, we perform single-domain TTA by adapting on the first domain and evaluating on the remaining domains. We see that across nearly all corruption categories, the adapted $\mathcal{W}_Q, \mathcal{W}_K, \mathcal{W}_V$ consistently outperform the SOURCE baseline (pre-trained). We see strong intra-category generalization; $\mathcal{W}_Q, \mathcal{W}_K, \mathcal{W}_V$ learned on \textit{Gaussian Noise} generalize effectively to \textit{Shot} and \textit{Impulse Noise}. This is consistent in other categories, too. We also notice cross-category transferability. In \textit{Blur}, the parameters learned on \textit{Gaussian Noise} perform slightly better than SOURCE, but are outperformed by the \textit{Blur}-specific parameters. Similar trends are in \textit{Digital}, too. This \textit{suggests} that the attention fusion layer captures domain-invariant correlations, enabling a \textit{reuse} of past parameters across related corruption categories.

We perform a similar evaluation on Kinetics50-2C, constructed by applying challenging bimodal audio-visual corruptions from \cite{maharana2025texttt} to the Kinetics50 test set. Following their taxonomy, corruptions are grouped into \textit{Digital}, \textit{Environmental}, and \textit{Human-Related} categories. We draw similar conclusions as earlier. Now, with the audio also being shifted, causing more data challenges, $\mathcal{W}_Q, \mathcal{W}_K, \mathcal{W}_V$ learned on \textit{Gaussian Noise} consistently outperform SOURCE throughout, while also generalizing really well to other domains within \textit{Digital}. We also observe patterns of cross-category transferability, as earlier.

\noindent \textit{\underline{Major takeaway.}} As observed, the adapted $\mathcal{W_Q}$, $\mathcal{W_K}$, and $\mathcal{W_V}$ of the joint-encoder $f_j$ on a single domain (for instance, \textit{Gaussian Noise}) yields strong intra-category and cross-category transfers, beating the SOURCE baseline. To exploit this transfer to unseen tasks/domains in a continual audio-visual setting, specifically, we propose using a shared buffer that can save model ``snapshots". This way, we can \textit{selectively} retrieve a previously optimized state and regain high performance. This leverages past successful adaptations to enable further, ongoing adaptation. When an input-level shift is detected, the model queries the buffer for the most ``relevant" parameters rather than continuing from a drifted state, as seen in the results of Figure \ref{fig:error}. Consequently, two critical design questions remain unanswered: 1) \textbf{Selective retrieval}: How do we identify and retrieve the optimal $\{\mathcal{W}_Q, \mathcal{W}_K, \mathcal{W}_V\}$ for adaptation at time-step $t$? 2) \textbf{Buffer expansion}: What criterion determines the addition of a new element to the shared buffer? Can we constrain the buffer size due to memory requirements?

\begin{figure*}[t!]
    \centering
    \includegraphics[trim={5mm 35mm 5mm 30mm },clip,width=\linewidth]{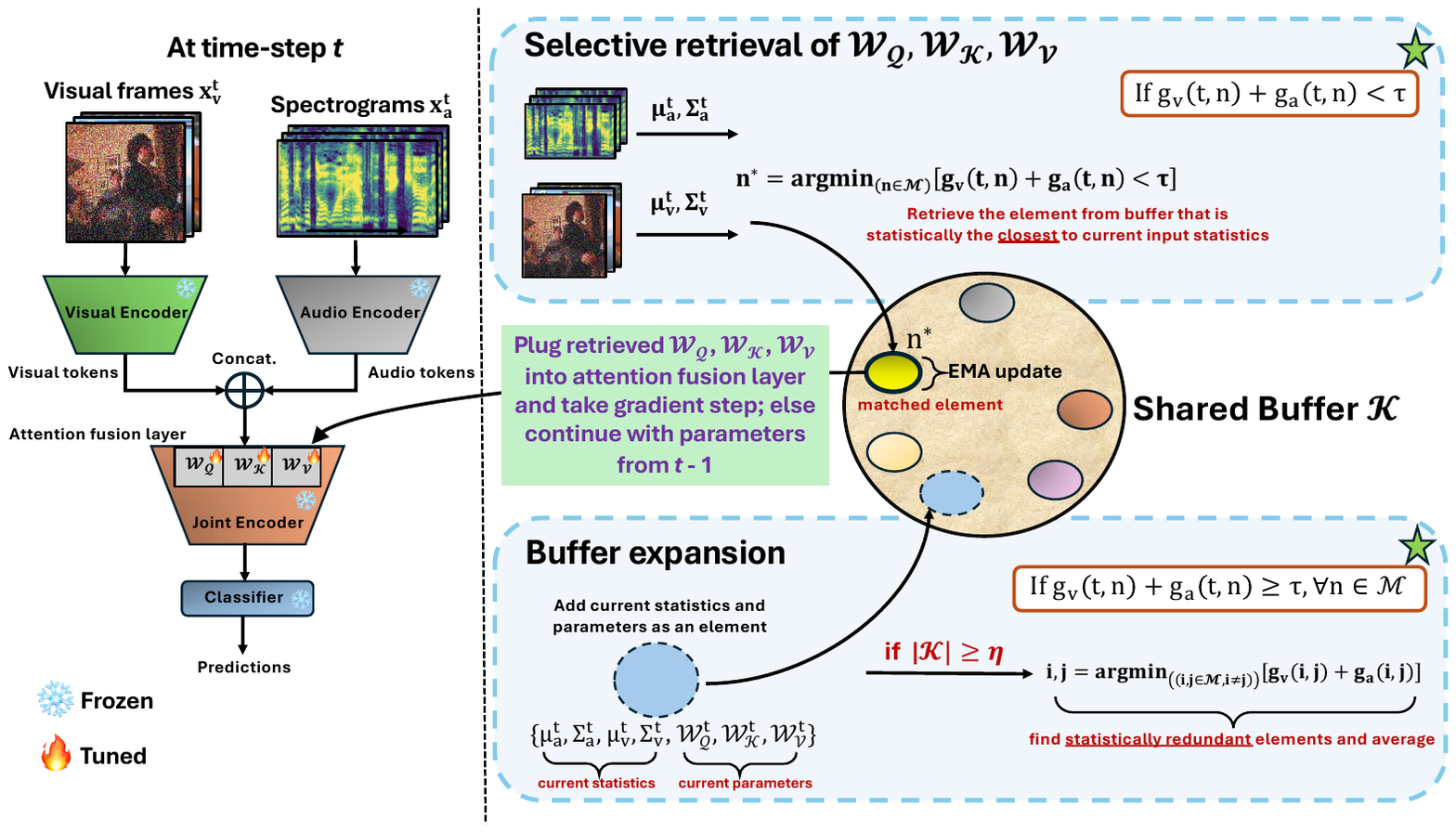}
    \caption{\textbf{Illustration of $\framework$.} At time-step $t$, audio-visual inputs via $\mu_a^t, \Sigma_a^t, \mu_v^t, \Sigma_v^t$ are modeled as Gaussian distributions. The \textbf{selection retrieval} stage uses KL divergence $g(\cdot)$ to compare current statistics against all elements in the shared buffer $\mathcal{K}$ ($\mathcal{M}$ is the current set of indices). For the best match within threshold $\tau$, stored parameters ($\mathcal{W}_Q, \mathcal{W}_K, \mathcal{W}_V$) are retrieved for adaptation at time-step $t$. If not, the \textbf{buffer expansion} stage involves adding current statistics and parameters in $\mathcal{K}$. Redundant elements are merged to maintain a memory budget $\eta$. Continual adaptation proceeds with current parameters, \ie from time-step $t-1$.}
    \label{fig:main}
    \vspace{-18pt}
\end{figure*}

We answer these questions below. In the Supplementary, we provide a neat algorithm of our proposed method. Figure \ref{fig:main} illustrates our approach.

\subsection{Selective retrieval of $\mathcal{W}_Q, \mathcal{W}_K, \mathcal{W}_V$}\label{method:retrieval}
In Figure \ref{fig:error}, we see that continual adaptation of $\mathcal{W}_Q, \mathcal{W}_K, \mathcal{W}_V$ in joint-encoder $f_j$ to bimodal shifts, following READ \cite{yang2024test}, leads to eventual overfitting. This is due to poor and degraded online cross-modal learning and fusion, with unrestricted continual parameter updates. In \S \ref{analysis}, we observed robust cross-domain generalization of the adapted attention fusion layer, outperforming the source model. To leverage this transferability during continual parameter updates, we propose utilizing a shared buffer that can save parameter snapshots, \ie $\mathcal{W}_Q, \mathcal{W}_K, \mathcal{W}_V$, to enable selective retrieval during audio-visual CTTA.

At \textit{t} = 0, \ie before adaptation begins with source model parameters $\theta^S$, the buffer $\mathcal{K}=\phi$. At time-step $t=$ 1 when the first set of inputs arrive ($x^1$), we follow READ and only online adapt $\mathcal{W}_Q, \mathcal{W}_K, \mathcal{W}_V$ where the loss function is described in \ref{optim} and $\theta^S$ $\rightarrow$ $\theta^1$. For $t>$ 1 and to enable selective retrieval of these parameters that can be plugged back and adapted, we characterize each $x^t$ by its underlying distribution. To note, the true distribution $p(x^t, y^t)$ is unavailable. Our selection criterion, instead, depends on the statistics (mean and covariance) of the raw data $x^t$ to capture modality-specific information via Gaussian distributions, owing to their simplicity and compactness. We assume a diagonal covariance structure for independence, \ie $\Sigma = \textrm{diag}(\sigma_1^2, \cdots,\sigma_r^2)$, where $r$ is the dimension. Let the visual frames be $x_v^t$ $\in$ $\mathbb{R}^{B \times H \times W \times C}$, where \textit{H}, \textit{W}, \textit{C} denote the height, width, and channel dimension, respectively. The corresponding spectrograms are $x_a^t$ $\in$ $\mathbb{R}^{B \times T \times F}$, where \textit{T} and \textit{F} denote the time and frequency axes, respectively. For the visual modality, we compute the channel-wise mean and covariance as, 

\begin{align}
\mu_v^t &= \frac{1}{BHW} \sum_{b=1}^B \sum_{h=1}^H \sum_{w=1}^W x_v^t(b, h, w), \label{stats_v} \\
\Sigma_v^t &= \frac{1}{BHW} \sum_{b=1}^B \sum_{h=1}^H \sum_{w=1}^W \left(x_v^t(b, h, w) - \mu_v^t\right)^2 \label{stats_v_cov}
\end{align}

Similarly, for the audio modality, we compute the mean and variance along the time axis $T$ to capture the frequency information of the spectrogram. That is, 
\begin{align}
\mu_a^t &= \frac{1}{BT} \sum_{b=1}^B \sum_{e=1}^T x_a^t(b,e), \label{stats_a} \\
\Sigma_a^t &= \frac{1}{BT} \sum_{b=1}^B \sum_{e=1}^T \left(x_a^t(b,e) - \mu_a^t\right)^2 \label{stats_a_cov}
\end{align}

Clearly, $\mu_v^t$, $\Sigma_v^t$ $\in$ $\mathbb{R}^C$ and $\mu_a^t$, $\Sigma_a^t$ $\in$ $\mathbb{R}^F$. At \textit{t} = 1, the buffer $\mathcal{K}$ is first initialized with an element consisting of $\{\mu_v^1, \Sigma_v^1, \mu_a^2, \Sigma_a^2, \mathcal{W}_Q^1, \mathcal{W}_K^1, \mathcal{W}_V^1\}$, where $\mathcal{W}_Q^1, \mathcal{W}_K^1, \mathcal{W}_V^1$ refer to the parameters \textit{after adaptation}. We assume that, throughout, the buffer takes the form $\mathcal{K} = \big[\{d_1, \widehat{\theta}_1\}, \{d_2, \widehat{\theta}_2\}, \{d_3, \widehat{\theta}_3\}, \ldots \big].$ Any element in $\mathcal{K}$ with index $n$ is $\mathcal{K}[n]$ = \{$d_n$, $\widehat\theta_n$\} where $d_n$ = $(\mu_v^n, \Sigma_v^n, \mu_a^n, \Sigma_a^n)$ and $\widehat\theta_n$ = $(\mathcal{W}_Q^n, \mathcal{W}_K^n, \mathcal{W}_V^n)$. $\mathcal{K}[n]$ could refer to any pair in $\mathcal{K}$ from a previous time-step since the domain boundaries are unknown in the continual setting, unlike \S \ref{analysis}. For clarification, an element \{$d_n$, $\widehat\theta_n$\} is only added as per our proposed \S \ref{method:buffer_}. We will discuss this shortly.

For $t>$ 1 with multimodal inputs $x_a^t$ and $x_v^t$ and corresponding input-statistics $(\mu_v^t, \Sigma_v^t, \mu_a^t, \Sigma_a^t)$, we perform a selective retrieval to identify the most compatible parameter state within the buffer. Our goal is to select the buffer element that is closest to the current input-statistics, retrieve its corresponding parameter state, and plug it back into the model for further adaptation. This way, as illustrated by Figure \ref{fig:attn_generalization}, we can possibly exploit transferability at any time-step when inputs arrive continually. This directly operationalizes the strong intra-category and cross-category generalizations, as observed earlier. For this, we model two modality-specific Gaussian distributions as $P_v^t = \mathcal{N}(\mu_v^t, \Sigma_v^t)$ and $P_a^t = \mathcal{N}(\mu_a^t, \Sigma_a^t)$. We define a distance metric $g_u(t, n)_{u\in\{a, v\}}$ (say, KL divergence), which is the distance between modality \textit{u}'s distribution at time-step \textit{t} and a stored buffer element $n$ (modeled as a corresponding distribution). Ideally, $g_u(t, n) = \mathcal{D}_{KL}(\mathcal{N}(\mu_u^t, \Sigma_u^t) \parallel \mathcal{N}(\mu_u^n, \Sigma_u^n))$ or $\mathcal{D}_{KL}^u$ as a short-hand. This degenerates to a closed-form solution (see the Supplementary). Now, we sum the modality-specific distances $g^{*}(t,n) = g_a(t,n)+g_v(t,n)$ to best capture the complete audio-visual input distance at time-step $t$. In the case of a unimodal corruption setting, let's say with visual-only corruptions, $g_a(t,n)$ might be negligible and $g_v(t,n)$ drives the retrieval. In a bimodal corruption setting, however, $g_v(t,n)$ and $g_a(t,n)$ both contribute.

Now, from here, two cases arise. For a threshold $\tau$, $g^{*}(t,n)<\tau$ or $g^{*}(t,n)\ge\tau$. $g^{*}(t,n)<\tau$ indicates that the $n^{th}$ element is distributionally closer to $x_a^t$ and $x_v^t$, \ie in the raw input space. As seen earlier in Figure \ref{fig:attn_generalization}, for example, attention fusion parameters adapted to \textit{Gaussian Noise} transferred to \textit{Shot Noise} and \textit{Impulse Noise} with improved performances over the source model. Hence, at any time-step $t$ in the continual setting, we are interested in finding the best or matched buffer element that is the closest to $x_a^t$ and $x_v^t$. To select, we solve the following optimization problem, 
\begin{align}\label{best_ele}
n^* = \underset{n \in \mathcal{M}}{\arg\!\min} \left[g^{*}(t,n)<\tau\right]
\end{align}
where $\mathcal{M}$ is the current set of indices in $\mathcal{K}$. The selected buffer element is $\mathcal{K}[n^*]$. Based on this, we retrieve the corresponding parameters $(\mathcal{W}_Q^{n^*}, \mathcal{W}_K^{n^*}, \mathcal{W}_V^{n^*})$. These selected parameters represent the configuration of the attention fusion layer projection matrices that most closely align with the current inputs $x^t$. We plug these specific parameters into $f_j$, indicating that the model had already seen a statistically similar audio-visual shift. In this way, from the raw input-space with domain shift, we exploit the intra-category and cross-category transferability. Following losses as proposed by READ (see \ref{optim}), we adapt the parameters once, \ie $\theta^{t-1} \rightarrow \theta^{t}$ where $\theta^t$ refers to $(\mathcal{W}_Q^{n^*}, \mathcal{W}_K^{n^*}, \mathcal{W}_V^{n^*})$ in joint-encoder $f_j$.

In addition to this, we perform a moment-preserving exponential moving average (EMA) update of the retrieved input-statistics $(\mu_v^{n^*}, \Sigma_v^{n^*}, \mu_a^{n^*}, \Sigma_a^{n^*})$ and parameter state $(\mathcal{W}_Q^{n^*}, \mathcal{W}_K^{n^*}, \mathcal{W}_V^{n^*})$. This ensures a smooth transition and encourages transferability, \ie intra-task and cross-task. Let $\beta$ (usually 0.99) be the EMA smoothing factor. For brevity, we denote $O \in \{Q, K, V \}$. Each parameter in the selected element of $\mathcal{K}$ is then updated as, 
\begin{align}\label{param_ema}
    \mathcal{W'}_O^{n^*} \leftarrow \beta \mathcal{W}_O^{n^*} + (1-\beta)\mathcal{W}_O^{t}
\end{align}
An EMA update of the input-statistics is done as follows, 
\begin{align}\label{input_ema}
\mu_u'^{n^*} &\leftarrow \beta \mu_u^{n^*} + (1-\beta)\mu_u^{t} \\
\Sigma_u'^{n^*} 
&\leftarrow \beta \left[ \Sigma_u^{n^*}  + (\mu_u^{n^*} - \mu_u'^{n^*})^2 \right] \notag \\
&\quad + (1-\beta) \left[ \Sigma_u^{t} + (\mu_u^{t} - \mu_u'^{n^*})^2 \right]\label{input_ema1}
\end{align}
where $\mu_u'^{n^*}$ and $\Sigma_u'^{n^*}$ denote updated modality-specific mean and diagonal covariance, and $u \in \{a,v\}$. Overall, the different components of the selected element are updated to $\mathcal{K}[n^*] \leftarrow \{\mu_v'^{n^*}, \Sigma_v'^{n^*}, \mu_a'^{n^*}, \Sigma_a'^{n^*}, \mathcal{W'}_Q^{n^*}, \mathcal{W'}_K^{n^*}, \mathcal{W'}_V^{n^*}\}$ via Eqns. \eqref{param_ema}, \eqref{input_ema}, and \eqref{input_ema1}. 

\subsection{Buffer expansion}\label{method:buffer_}
In our method, an element is added to $\mathcal{K}$ \textit{only when} $g^{*}(t,n)\ge\tau$, $\forall n \in \mathcal{M}$. This means that all the current elements in $\mathcal{K}$ are distributionally distinct from $x_a^t$ and $x_v^t$. Since no parameters were retrieved, we optimize the current model parameters as-is, following \ref{optim}. Once a gradient step is taken, we add a new element to $\mathcal{K}$ as $\mathcal{K}$ $\cup$ $\{\mu_v^t, \Sigma_v^t, \mu_a^t, \Sigma_a^t, \mathcal{W}_Q^t, \mathcal{W}_K^t, \mathcal{W}_V^t\}$. This new element will serve as an anchor for future occurrences. A possibility in such a scenario is the unconstrained growth of the buffer, a key consideration in memory-constrained continual learning \cite{smith2021memory}. To address this, we explore a variant of \textit{element merging}, \ie for a fixed buffer budget/size $\eta$, we perform a pairwise comparison (when $|\mathcal{K}| \ge \eta$) to find out the two nearest elements based on their statistics and merge them. That is, for indices $n_1$ and $n_2$, we solve,
\begin{align}\label{merge_eq}
(n_1, n_2) = \underset{i, j \in \mathcal{M}, i \neq j}{\arg\!\min} g^{*}(i,j) 
\end{align}
Once the most similar pair $(n_1, n_2)$ is identified, we merge them to maintain the fixed budget $\eta$. As an approximation, we average the corresponding elements in $\mathcal{K}[n_1]$ and $\mathcal{K}[n_2]$ and add the element back in $\mathcal{K}$. Now, $|\mathcal{K}| = \eta - 1$. We show an ablation in \ref{ablation_eta} and explore another strategy in the Supplementary.

\subsection{Losses}\label{optim}
At every time-step $t$, we optimize following the loss functions proposed in READ \cite{yang2024test}. For a notational purpose, let $\sigma$ denote the softmax operation. Let the prediction confidence of the $i^{th}$ input be $p_{\max} = \max(\sigma(l_i))$, where $l_i$ refers to the logits from the classifier. For a batch of $B$ inputs, the confidence-based loss function $\mathcal{L}_{conf}$ is, 
\begin{align}
   \mathcal{L}_{conf} = \mathbb{E}_{i\sim B}[-p_{i,max}\mathrm{log}(p_{i,max})] 
\end{align}
In addition, a negative entropy loss \cite{li2024comprehensive} as below is also used, 
\begin{align}
    \mathcal{L}_{ne} = \sum_{a=1}^{A}\sigma(k_a)\mathrm{log}(\sigma(k_a))
\end{align}
where, $k_a$ = $\sum_{i=1}^B \sigma(l_i)$. $A$ refers to the specific class. The complete loss function is, $\mathcal{L}$ = $\mathcal{L}_{conf}$ + $\mathcal{L}_{ne}$, where only $\mathcal{W}_Q, \mathcal{W}_K, \mathcal{W}_V$ of the the joint-encoder $f_j$ are \textbf{continually} adapted.

\subsection{Motivation of this design choice}
In \cite{yang2024test}, the authors identified the attention fusion layer, with frozen encoders, as critical for cross-modal TTA. As a follow-up, \cite{maharana2024texttt} showed that bimodal corruptions induce attention imbalance, i.e, in online TTA, there is a widening gap in self and cross-attention between the visual and audio tokens. Motivated by
these observations, our method focuses on adapting only the fusion parameters and retrieves similar parameters from the shared buffer instead of continually overwriting them. Throughout, both encoders are frozen. We later show that this design choice, combined with our proposed method $\framework$, minimizes catastrophic forgetting (Figure \ref{fig:cf}).

\section{Experiments}
\label{results}

\noindent \textbf{Settings.} We evaluate in two main corruption settings - unimodal and bimodal. Corruptions are applied to either one modality or both, respectively. Here, tasks arrive sequentially (see \S \ref{setting}), and parameter updates happen continually without any reset. Each corruption spans the entire test set and defines a task.

\noindent \textbf{Datasets.} We evaluate on the test sets of Kinetics50 \cite{carreira2017quo} and VGGSound \cite{chen2020vggsound}, following READ \cite{yang2024test}. For unimodal corruptions, we use Kinetics50-C and VGGSound-C, which include 15 visual \cite{hendrycks2019benchmarking} and 6 audio corruptions at severity level 5. For bimodal corruptions, we adopt 15 corruptions from AVRobustBench \cite{maharana2025texttt}, applied to both modalities, resulting in Kinetics50-2C and VGGSound-2C. Kinetics50 is \textit{visually dominant}, while VGGSound is \textit{audio dominant}, meaning task-relevant cues primarily reside in the visual and audio modalities, respectively. More details are provided in the Supplementary.

\noindent \textbf{Baselines.} We compare our proposed $\framework$ to popular TTA baselines (extensible to continual) like TENT \cite{wang2020tent}, EATA \cite{niu2022efficient}, and SAR \cite{niu2023towards}. The prior audio-visual baselines include READ \cite{yang2024test}, SuMi \cite{guo2025smoothing}, PTA \cite{wangpartition}, and BriMPR* \cite{li2025bridging}. As discussed in \S \ref{remark}, we eliminate access to the source data in BriMPR to make it more deployment-friendly in a real-world setting. We also do not compare against \cite{zhang2025analytic} as it requires full access to source data.

\noindent \textbf{Implementation Details.} Following previous works, we use CAV-MAE \cite{gong2022contrastive} as the source model, trained on Kinetics50 or VGGSound. We use a batch size of 32 across all experiments (see Supplementary for an ablation). We optimize with Adam using a learning rate of $1\times10^{-4}$. For unimodal corruptions, we set $\tau$ to 0.005 and 0.01 for bimodal corruptions. For all the baselines, we adopt their recommended hyperparameters. We expand on the details in the Supplementary. All experiments are conducted on an NVIDIA RTX A5000 GPU.

\begin{table*}[t!]
\centering
\setlength{\tabcolsep}{3pt}
\scriptsize
\caption{\textbf{$\framework$ achieves SOTA results on Kinetics50-C} (top) \textbf{and VGGSound-C} (below) with \textbf{video corruptions} (unimodal). We report the task-wise accuracy (\%) at a severity level of 5, in the \textbf{continual} setting.}
\begin{adjustbox}{width=\textwidth}
\begin{tabular}{cccccccccccccccccc}
\toprule
\textbf{} & \textbf{Method} &
\multicolumn{3}{c}{\textbf{Noise}} &
\multicolumn{4}{c}{\textbf{Blur}} &
\multicolumn{4}{c}{\textbf{Weather}} &
\multicolumn{4}{c}{\textbf{Digital}} &
\textbf{Mean} \\
\cmidrule(lr){3-5} \cmidrule(lr){6-9} \cmidrule(lr){10-13} \cmidrule(lr){14-17}
\textbf{} & &
Gaussian & Shot & Impulse &
Defocus & Glass & Motion & Zoom &
Snow & Frost & Fog & Brightness &
Contrast & \shortstack{Elastic\\Transform} & Pixelate & JPEG &
\\
\midrule
\multirow{10}{*}{\rotatebox[origin=c]{90}{\normalsize Kinetics50-C}} &
SOURCE &
46.55 & 47.36 & 46.51 & 67.23 & 61.70 & 70.67 & 66.39 & 62.22 & 60.54 & 47.32 & 74.88 & 51.36 & 64.14 & 66.39 & 61.98 & 59.68
\\
& TENT &
45.55 & 42.51 & 27.84 &
22.76 & 8.33 & 3.64 & 2.24 &
2.40 & 2.08 & 2.00 & 2.04 &
2.28 & 2.04 & 2.00 & 2.00 &
11.31 \\
& EATA &
46.80 & 47.62 & 46.81 & 67.62 & 63.02 & 70.77 & 66.94 & 60.65 & 60.65 & 48.23 & 75.48 & 52.42 & 65.40 & 66.45 & 62.22 & 60.07
\\
& SAR &
46.83 & 47.00 & 46.64 & 65.71 & 62.30 & 69.99 & 66.35 & 60.54 & 60.22 & 52.60 & 73.96 & 50.76 & 65.99 & 63.42 & 59.90 & 59.48
\\
& READ &
49.96 & 51.80 & 51.56 & 68.07 & 65.63 & 65.38 & 61.50 & 52.20 & 49.48 & 42.19 & 54.93 & 28.93 & 39.82 & 33.17 & 27.32 & 49.46
\\
& SuMi &
45.75 & 45.43 & 44.95 & 64.30 & 65.38 & 69.23 & 67.91 & 62.86 & 66.67 & 63.22 & 69.23 & 50.16 & 71.15 & 66.31 & 63.94 & 61.10
\\
& PTA &
46.84 & 45.95 & 45.91 & 56.17 & 55.73 & 55.49 & 54.81 & 51.80 & 52.60 & 50.24 & 54.73 & 49.80 & 53.32 & 52.24 & 51.12 & 51.78
\\
& BriMPR* &
49.12 & 49.48 & 48.64 & 58.25 & 59.82 & 58.57 & 58.53 & 53.69 & 55.49 & 54.33 & 59.13 & 50.32 & 59.98 & 57.77 & 57.61 & 55.38
\\
& $\framework$ ($\eta=50$) &
\cellcolor[HTML]{C6F6D5}48.88 & \cellcolor[HTML]{C6F6D5}51.00 & \cellcolor[HTML]{C6F6D5}48.72 &
\cellcolor[HTML]{C6F6D5}68.23 & \cellcolor[HTML]{C6F6D5}64.78 & \cellcolor[HTML]{C6F6D5}70.99 & \cellcolor[HTML]{C6F6D5}68.47 &
\cellcolor[HTML]{C6F6D5}65.02 & \cellcolor[HTML]{C6F6D5}66.35 & \cellcolor[HTML]{C6F6D5}61.58 & \cellcolor[HTML]{C6F6D5}72.52 &
\cellcolor[HTML]{C6F6D5}53.81 & \cellcolor[HTML]{C6F6D5}66.79 & \cellcolor[HTML]{C6F6D5}68.19 & \cellcolor[HTML]{C6F6D5}65.99 &
\cellcolor[HTML]{C6F6D5}\textbf{62.75}
\\
& $\framework$ ($\eta=100$) &
\cellcolor[HTML]{C6F6D5}48.52 & \cellcolor[HTML]{C6F6D5}50.52 & \cellcolor[HTML]{C6F6D5}48.08 &
\cellcolor[HTML]{C6F6D5}67.95 & \cellcolor[HTML]{C6F6D5}63.86 & \cellcolor[HTML]{C6F6D5}71.59 & \cellcolor[HTML]{C6F6D5}67.75 &
\cellcolor[HTML]{C6F6D5}65.14 & \cellcolor[HTML]{C6F6D5}67.19 & \cellcolor[HTML]{C6F6D5}62.14 & \cellcolor[HTML]{C6F6D5}72.80 &
\cellcolor[HTML]{C6F6D5}54.17 & \cellcolor[HTML]{C6F6D5}66.43 & \cellcolor[HTML]{C6F6D5}67.59 & \cellcolor[HTML]{C6F6D5}65.18 &
\cellcolor[HTML]{C6F6D5}\textbf{62.59} \\
& $\framework$ ($\eta=\infty$) &
\cellcolor[HTML]{C6F6D5}48.52 & \cellcolor[HTML]{C6F6D5}50.24 & \cellcolor[HTML]{C6F6D5}48.32 &
\cellcolor[HTML]{C6F6D5}67.91 & \cellcolor[HTML]{C6F6D5}63.18 & \cellcolor[HTML]{C6F6D5}71.60 & \cellcolor[HTML]{C6F6D5}68.03 &
\cellcolor[HTML]{C6F6D5}64.86 & \cellcolor[HTML]{C6F6D5}67.15 & \cellcolor[HTML]{C6F6D5}62.10 & \cellcolor[HTML]{C6F6D5}72.80 &
\cellcolor[HTML]{C6F6D5}53.57 & \cellcolor[HTML]{C6F6D5}65.43 & \cellcolor[HTML]{C6F6D5}67.87 & \cellcolor[HTML]{C6F6D5}64.02 &
\cellcolor[HTML]{C6F6D5}\textbf{62.37}
\\
\bottomrule

\multirow{10}{*}{\rotatebox[origin=c]{90}{\normalsize VGGSound-C}} &
SOURCE &
52.78 & 52.69 & 52.73 & 57.19 & 57.21 & 58.52 & 57.61 & 56.25 & 56.58 & 55.31 & 58.95 & 53.66 & 56.95 & 55.81 & 56.89 & 55.94
\\
& TENT &
52.58 & 51.82 & 51.20 & 53.85 & 53.90 & 54.49 & 54.39 & 51.80 & 52.39 & 52.59 & 52.14 & 51.08 & 52.49 & 51.70 & 51.87 & 52.55
\\
& EATA &
52.79 & 52.81 & 52.46 & 56.60 & 55.90 & 57.43 & 56.16 & 54.36 & 54.54 & 53.35 & 57.06 & 51.94 & 54.56 & 53.23 & 54.51 & 54.51
\\
& SAR &
52.92 & 52.86 & 52.92 & 57.03 & 56.66 & 58.65 & 57.61 & 55.95 & 56.74 & 56.22 & 58.25 & 54.21 & 57.13 & 55.42 & 56.46 & 55.94
\\
& READ &
52.43 & 52.65 & 52.44 & 55.85 & 55.22 & 56.07 & 55.69 & 54.51 & 54.96 & 54.69 & 55.38 & 54.34 & 54.79 & 54.65 & 54.65 & 54.55
\\
& SuMi & 
53.14 & 53.42 & 53.43 & 57.01 & 56.69 & 57.67 & 56.91 & 55.60 & 56.22 & 51.73 & 56.91 & 53.67 & 56.37 & 55.16 & 55.25 & 55.28\\
& PTA &
51.06 & 49.86 & 49.46 & 52.88 & 52.12 & 52.31 & 51.94 & 50.96 & 51.21 & 51.17 & 51.92 & 50.46 & 51.59 & 51.03 & 51.05 & 51.27
\\
& BriMPR* &
43.24 & 39.11 & 38.86 & 39.99 & 41.22 & 39.41 & 37.97 & 33.12 & 30.76 & 31.47 & 30.71 & 26.08 & 31.95 & 34.30 & 36.15 & 35.62
\\

& $\framework$ ($\eta=300$) &
\cellcolor[HTML]{C6F6D5}52.87 & \cellcolor[HTML]{C6F6D5}52.87 & \cellcolor[HTML]{C6F6D5}52.86 & \cellcolor[HTML]{C6F6D5}57.16 & \cellcolor[HTML]{C6F6D5}56.97 & \cellcolor[HTML]{C6F6D5}58.65 & \cellcolor[HTML]{C6F6D5}58.02 & \cellcolor[HTML]{C6F6D5}56.66 & \cellcolor[HTML]{C6F6D5}57.13 & \cellcolor[HTML]{C6F6D5}56.67 & \cellcolor[HTML]{C6F6D5}58.18 & \cellcolor[HTML]{C6F6D5}54.66 & \cellcolor[HTML]{C6F6D5}57.42 & \cellcolor[HTML]{C6F6D5}56.22 & \cellcolor[HTML]{C6F6D5}56.71 & \cellcolor[HTML]{C6F6D5}\textbf{56.20} \\
& $\framework$ ($\eta=400$) &
\cellcolor[HTML]{C6F6D5}52.86 & \cellcolor[HTML]{C6F6D5}52.83 & \cellcolor[HTML]{C6F6D5}52.91 & \cellcolor[HTML]{C6F6D5}57.30 & \cellcolor[HTML]{C6F6D5}57.25 & \cellcolor[HTML]{C6F6D5}58.66 & \cellcolor[HTML]{C6F6D5}57.95 & \cellcolor[HTML]{C6F6D5}56.50 & \cellcolor[HTML]{C6F6D5}56.76 & \cellcolor[HTML]{C6F6D5}56.47 & \cellcolor[HTML]{C6F6D5}57.89 & \cellcolor[HTML]{C6F6D5}54.80 & \cellcolor[HTML]{C6F6D5}57.40 & \cellcolor[HTML]{C6F6D5}55.82 & \cellcolor[HTML]{C6F6D5}56.54 & \cellcolor[HTML]{C6F6D5}\textbf{56.13} \\
& $\framework$ ($\eta=\infty$) &
\cellcolor[HTML]{C6F6D5}52.87 & \cellcolor[HTML]{C6F6D5}52.83 & \cellcolor[HTML]{C6F6D5}52.91 &
\cellcolor[HTML]{C6F6D5}57.42 & \cellcolor[HTML]{C6F6D5}57.29 & \cellcolor[HTML]{C6F6D5}58.68 & \cellcolor[HTML]{C6F6D5}57.77 &
\cellcolor[HTML]{C6F6D5}56.38 & \cellcolor[HTML]{C6F6D5}56.80 & \cellcolor[HTML]{C6F6D5}55.50 & \cellcolor[HTML]{C6F6D5}59.13 &
\cellcolor[HTML]{C6F6D5}53.88 & \cellcolor[HTML]{C6F6D5}57.07 & \cellcolor[HTML]{C6F6D5}55.96 & \cellcolor[HTML]{C6F6D5}57.01 &
\cellcolor[HTML]{C6F6D5}\textbf{56.10}
\\
\bottomrule
\end{tabular}
\end{adjustbox}
\label{tab:uni_video}
\end{table*}

\begin{table}[htb!]
\centering
\setlength{\tabcolsep}{2.5pt}
\scriptsize
\caption{\textbf{$\framework$ achieves SOTA results on Kinetics50-C} (top) \textbf{and VGGSound-C} (below) with \textbf{audio corruptions} (unimodal). We report the task-wise accuracy (\%) at a severity level of 5, in the \textbf{continual} setting.}
\begin{adjustbox}{width=0.5\textwidth}
\begin{tabular}{ccccccccc}
\toprule
\textbf{} & \textbf{Method} &
\multicolumn{3}{c}{\textbf{Noise}} &
\multicolumn{3}{c}{\textbf{Weather}} &
\textbf{Mean} \\
\cmidrule(lr){3-5} \cmidrule(lr){6-8} 
\textbf{} & &
Gaussian & Traffic & Crowd &
Rain & Thunder & Wind &
\\
\midrule
\multirow{10}{*}{\rotatebox[origin=c]{90}{\normalsize Kinetics50-C}} &
SOURCE &
73.48 & 65.30 & 67.59 & 70.11 & 67.67 & 70.11 & 69.04 
\\
& TENT &
73.84 & 68.55 & 70.51 & 69.59 & 73.12 & 70.15 & 70.96 \\
& EATA &
73.56 & 65.30 & 67.71 & 70.11 & 68.07 & 70.15 & 69.15
\\
& SAR &
73.36 & 65.95 & 68.11 & 69.71 & 69.15 & 69.71 & 69.33
\\
& READ &
74.28 & 69.63 & 70.63 & 70.35 & 71.76 & 69.39 & 71.01
\\
& SuMi &
73.76 & 68.19 & 70.59 & 69.27 & 73.24 & 69.43 & 70.75
\\
& PTA &
72.68 & 69.03 & 69.79 & 68.83 & 71.79 & 69.75 & 70.31
\\
& BriMPR* &
72.84 & 67.67 & 67.43 & 65.95 & 69.07 & 63.70 & 67.78
\\
& $\framework$ ($\eta=50$) &\
\cellcolor[HTML]{C6F6D5}73.72 & \cellcolor[HTML]{C6F6D5}68.35 & \cellcolor[HTML]{C6F6D5}70.27 &
\cellcolor[HTML]{C6F6D5}70.11 & \cellcolor[HTML]{C6F6D5}72.88 & \cellcolor[HTML]{C6F6D5}69.91 &
\cellcolor[HTML]{C6F6D5}70.87 \\
& $\framework$ ($\eta=100$) &\
\cellcolor[HTML]{C6F6D5}73.72 & \cellcolor[HTML]{C6F6D5}68.47 & \cellcolor[HTML]{C6F6D5}69.79 &
\cellcolor[HTML]{C6F6D5}70.27 & \cellcolor[HTML]{C6F6D5}72.56 & \cellcolor[HTML]{C6F6D5}70.39 &
\cellcolor[HTML]{C6F6D5}70.87  \\
& $\framework$ ($\eta=\infty$) &
\cellcolor[HTML]{C6F6D5}74.15 & \cellcolor[HTML]{C6F6D5}68.90 & \cellcolor[HTML]{C6F6D5}70.30 &
\cellcolor[HTML]{C6F6D5}70.62 & \cellcolor[HTML]{C6F6D5}72.06 & \cellcolor[HTML]{C6F6D5}70.57 &
\cellcolor[HTML]{C6F6D5}\textbf{71.10}
\\
\bottomrule

\multirow{10}{*}{\rotatebox[origin=c]{90}{\normalsize  VGGSound-C}} &
SOURCE &
37.36 & 21.12 & 16.80 & 21.64 & 27.29 & 25.54 & 24.96
\\
& TENT &
6.20 & 0.46 & 0.29 & 0.28 & 0.28 & 0.28 & 1.30
\\
& EATA &
37.51 & 21.64 & 17.58 & 22.82 & 27.98 & 25.34 & 25.48
\\
& SAR &
36.53 & 8.01 & 4.12 & 4.49 & 13.43 & 3.34 & 11.65
\\
& READ &
28.01 & 15.10 & 17.35 & 13.69 & 20.24 & 14.37 & 18.13
\\
& SuMi & 
37.66 & 19.28 & 14.79 & 20.73 & 28.82 & 28.05 & 24.89\\
& PTA &
36.30 & 28.79 & 28.42 & 25.35 & 30.60 & 26.09 & 29.26
\\
& BriMPR* &
23.12 & 14.95 & 15.15 & 13.85 & 20.08 & 15.24 & 17.07
\\
& $\framework$ ($\eta=300$) &
\cellcolor[HTML]{C6F6D5}39.52 & \cellcolor[HTML]{C6F6D5}28.43 & \cellcolor[HTML]{C6F6D5}24.36 &
\cellcolor[HTML]{C6F6D5}29.81 & \cellcolor[HTML]{C6F6D5}33.58 & \cellcolor[HTML]{C6F6D5}20.85 &
\cellcolor[HTML]{C6F6D5}\textbf{29.42} \\
& $\framework$ ($\eta=400$) &
\cellcolor[HTML]{C6F6D5}39.52 & \cellcolor[HTML]{C6F6D5}28.43 & \cellcolor[HTML]{C6F6D5}24.32 &
\cellcolor[HTML]{C6F6D5}29.37 & \cellcolor[HTML]{C6F6D5}35.65 & \cellcolor[HTML]{C6F6D5}26.38 &
\cellcolor[HTML]{C6F6D5}\textbf{30.61}\\
& $\framework$ ($\eta=\infty$) &
\cellcolor[HTML]{C6F6D5}39.52 & \cellcolor[HTML]{C6F6D5}28.43 & \cellcolor[HTML]{C6F6D5}24.31 &
\cellcolor[HTML]{C6F6D5}29.44 & \cellcolor[HTML]{C6F6D5}35.52 & \cellcolor[HTML]{C6F6D5}27.05 &
\cellcolor[HTML]{C6F6D5}\textbf{30.71}
\\
\bottomrule
\end{tabular}
\end{adjustbox}
\label{tab:uni_audio}
\vspace{-10pt}
\end{table}

\subsection{Main Results}

\noindent \textbf{Results on the unimodal corruption setting.} In Tables \ref{tab:uni_video} and \ref{tab:uni_audio}, we present results for unimodal video and audio corruptions, respectively. Under visual corruptions in Table \ref{tab:uni_video}, particularly on Kinetics50-C, where task information is heavily degraded, TENT exhibits severe overfitting, as entropy minimization updates all LayerNorm \cite{ba2016layer} affine parameters across attention blocks in all encoders. This update is independent of any modality interaction or cross-modal fusion. In contrast, EATA and SAR also update norm parameters but maintain performance due to their respective loss functions, achieving accuracy comparable to the source model. In READ, continual updates to $\mathcal{W_Q}, \mathcal{W_V}, \mathcal{W_V}$ lead to an eventual decline, as continual adaptation amplifies modality imbalance \cite{maharana2025texttt} and error accumulation, ultimately biasing the attention mechanism. Without access to source data, the randomly initialized visual prompts in BriMPR* struggle to maintain semantic alignment and fail to effectively adapt to the target domains. At a small budget $\eta=50$, $\framework$ achieves a relative improvement of 2.7\% over the next best method, SuMi, on average. On a larger test set like VGGSound-C with audio corruptions, we have similar observations and $\framework$ obtains a relative improvement of 4.95\% over PTA. On VGGSound-C with visual corruptions and Kinetics50-C with audio corruptions, \ie where task-dominant information is clean, $\framework$ achieves comparable to better mean accuracies. We show more results with varying $\eta$ in \ref{ablation_eta} (ablation), and in the Supplementary.

\begin{figure}[t!]
    \centering
    \begin{subfigure}{0.48\linewidth}
        \centering
        \includegraphics[width=\linewidth,trim={0mm 0mm 0mm 0mm},clip]{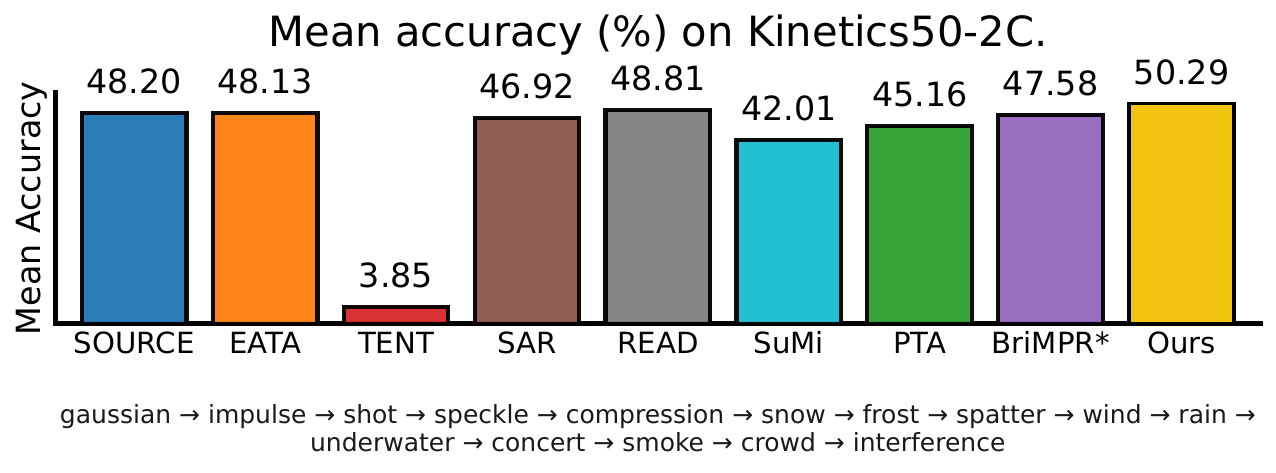}
        \caption{Kinetics50-2C}
        \label{fig:ks-2c}
    \end{subfigure}
    \hfill
    \begin{subfigure}{0.48\linewidth}
        \centering
        \includegraphics[width=\linewidth,trim={0mm 0mm 0mm 0mm},clip]{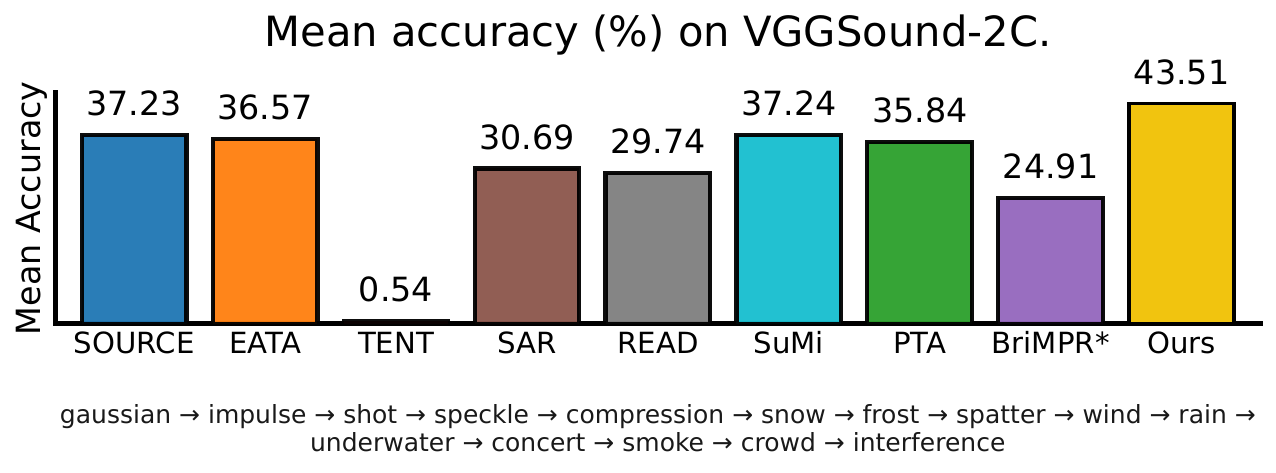}
        \caption{VGGSound-2C}
        \label{fig:vgg-2c}
    \end{subfigure}
    \caption{\textbf{$\framework$ achieves SOTA results on Kinetics50-2C} (left) \textbf{and VGGSound-2C} (right). We report mean accuracy (\%) at a severity level of 5 in the \textbf{continual} setting. Here, buffer size $\eta=\infty$.}
    \vspace{-15pt}
\end{figure}

\noindent \textbf{Results on the bimodal corruption setting.} In Figures \ref{fig:ks-2c} and \ref{fig:vgg-2c}, we illustrate the mean results in the challenging bimodal corruptions. In this, the task-wise accuracies are in the Supplementary. First off, TENT overfits drastically due to continual norm updates across domains. On Kinetics50-2C, the proposed $\framework$ obtains an improvement of 1.25\% over READ (next best). Interestingly, most methods fall short of a pre-trained SOURCE model under continual parameter updates with bimodal corruptions since cross-modal learning is hampered. A similar trend is observed on the more challenging VGGSound-2C benchmark with 309 classes and 14,046 samples in each task, where $\framework$ achieves a substantial 6.28\% improvement over SOURCE. In contrast, most methods progressively overfit and perform worse than SOURCE. The ablation results on buffer budget $\eta$ are in the Supplementary.

\subsection{Ablation Studies}

\begin{figure*}[t!]
\centering

\begin{minipage}[t]{0.49\textwidth}
    \centering
    \includegraphics[width=\linewidth]
    {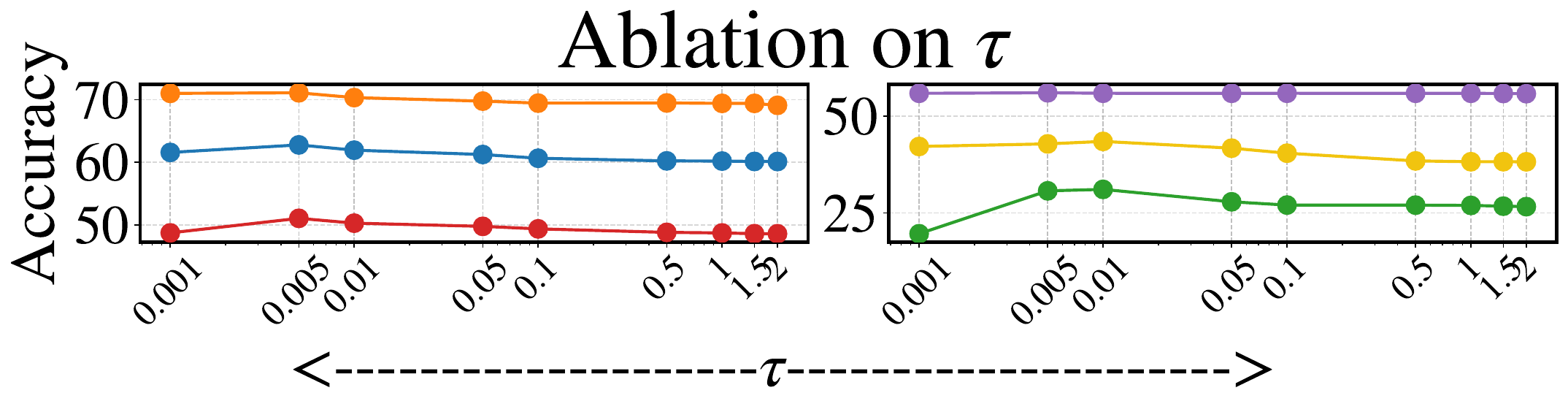}
    \caption{\textbf{Ablation on $\tau$.}
    Results on Kinetics50 (left) under \textcolor{Orange}{\textbf{audio}}, 
    \textcolor{RoyalBlue}{\textbf{visual}}, and 
    \textcolor{Red}{\textbf{bimodal}} corruptions, and on VGGSound (right) under 
    \textcolor{Plum}{\textbf{visual}}, 
    \textcolor{Goldenrod}{\textbf{audio}}, and 
    \textcolor{Green}{\textbf{bimodal}} corruptions.}
    \label{fig:tau_ablation}
\end{minipage}
\hfill
\begin{minipage}[t]{0.49\textwidth}
    \centering
    \includegraphics[trim={8 7 5 5},clip,width=0.7\linewidth]
    {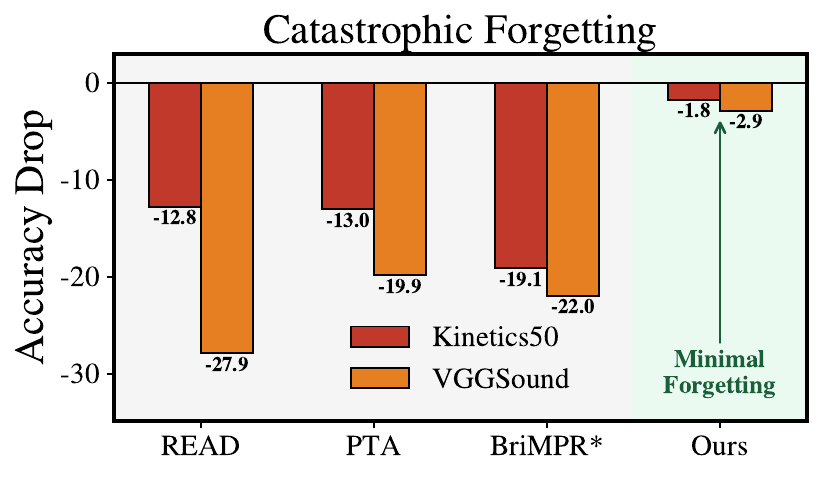}
    \caption{\textbf{$\framework$ minimizes \textit{catastrophic forgetting}.}
    We indicate the accuracy drop after continual adaptation to 
    Kinetics50-2C and VGGSound-2C compared to the respective source data 
    performance of CAV-MAE.}
    \label{fig:cf}
\end{minipage}
\vspace{-10pt}
\end{figure*}

\noindent \textbf{Effect of threshold $\tau$.} In Figure \ref{fig:tau_ablation}, we ablate the distance threshold $\tau$ and report results on both datasets. We sweep $\tau \in \{0.001, 0.005, 0.01, 0.05, 0.1, 0.5, 1, 1.5, 2\}$. Performance gradually degrades as $\tau$ increases, with a more pronounced drop under bimodal corruptions. Larger $\tau$ leads to larger buffers that allow minor distributional fluctuations to be added, as elements that reduce cross-domain transferability. In contrast, a smaller $\tau$ value strikes a favorable balance.

\noindent \textbf{Catastrophic forgetting.}\label{forget} We evaluate the final adapted parameters back on the source test set. We continually adapt on Kinetics50-2C and VGGSound-2C, and illustrate the results in Figure \ref{fig:cf}. For Kinetics50-2C, this corresponds to $15\times2466$ updates and $15\times14046$ updates for VGGSound-2C. As seen, in the presence of challenging bimodal corruptions, selective parameter retrieval, as done in our work, significantly preserves the source knowledge. In prior works, long-term continual parameter adaptation can lead to major source knowledge forgetting, as in Figure \ref{fig:cf}.

\begin{figure}[t!]
    \centering
    \begin{subfigure}{0.48\linewidth}
        \centering
        \includegraphics[width=0.8\linewidth,trim={6mm 5mm 6mm 5mm},clip]{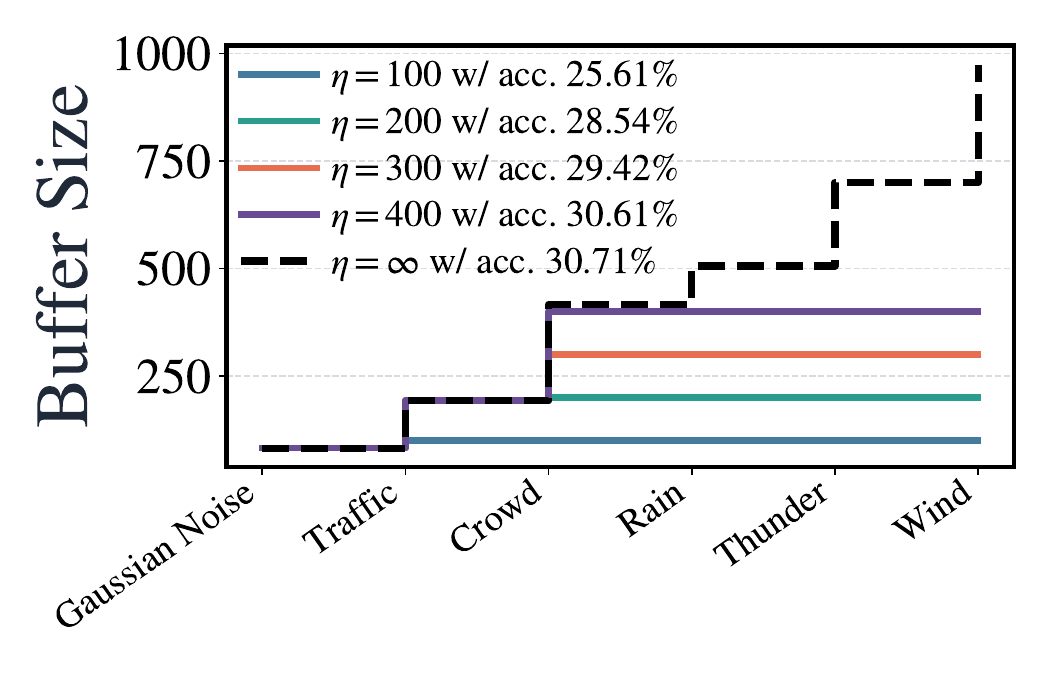}
        \caption{VGGSound-C w/ audio corruptions.}
        \label{fig:buffer-audio}
    \end{subfigure}
    \hfill
    \begin{subfigure}{0.48\linewidth}
        \centering
        \includegraphics[width=0.8\linewidth,trim={6mm 5mm 6mm 5mm},clip]{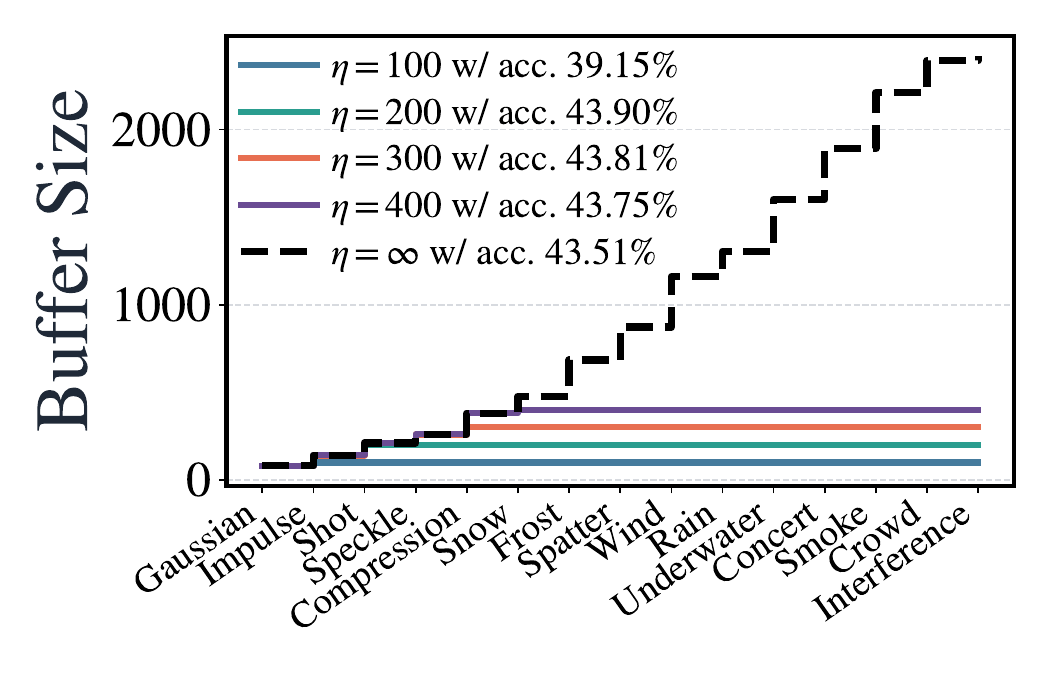}
        \caption{VGGSound-2C w/ bimodal corruptions.}
        \label{fig:buffer-video}
    \end{subfigure}
    \caption{\textbf{Effect of buffer size $\eta$.} Mean accuracy (\%) on VGGSound-C (left) and VGGSound-2C (right).}
    \label{fig:buffer}
    \vspace{-15pt}
\end{figure}

\noindent \textbf{Buffer budget $\eta$.}\label{ablation_eta}
In Figure \ref{fig:buffer}, we study the scalability of $\framework$ with a fixed buffer size $\eta$ on VGGSound-C (w/ audio corruptions) and VGGSound-2C. Each task comprises approximately 439 batches ($\frac{14,046}{32}$) across 6/15 tasks, respectively. For VGGSound-C, we notice that at a fixed budget $\eta\ge300$ throughout, $\framework$ still achieves better or competitive accuracies compared to baselines. On VGGSound-2C, with a tighter budget of $\eta=200$, we notice improved performances.  Element merging indeed consolidates task-specific knowledge and remains stable in long-term continual adaptation.

\begin{wraptable}{r}{0.48\columnwidth}
\vspace{-20pt}
\centering
\setlength{\tabcolsep}{3pt}
\scriptsize
\caption{\textbf{Alternative distance metric $g_u(t,n)$.}
Results on Kinetics50-C (visual corruptions) and Kinetics50-2C.}
\begin{adjustbox}{width=0.4\textwidth}
\begin{tabular}{lcc}
\toprule
\textbf{Metric $g_u(t,n)$} &
\textbf{K50-C} & \textbf{K50-2C} \\
\midrule
$\|\mu_u^t-\mu_u^n\|_2 + \|\sigma_u^t-\sigma_u^n\|_2$
& 49.47 & 48.79 \\
$\mathcal{D}_{KL}^{v}$ & 61.53 & 49.28 \\
$\mathcal{D}_{KL}^{a}$ & 60.52 & 49.70 \\
\rowcolor[HTML]{C6F6D5}
$\mathcal{D}_{KL}^{a+v}$ (Ours) & \textbf{62.37} & \textbf{50.29} \\
\bottomrule
\end{tabular}
\end{adjustbox}
\vspace{-15pt}
\label{tab:distance}
\end{wraptable}

\noindent \textbf{Distance metric $g_u(t,n)$.} In Table \ref{tab:distance}, we experiment with a norm-based distance as $g_u(t,n)=\| \mu_u^t - \mu_u^n\|_2$ +  $\| \sigma_u^t - \sigma_u^n\|_2$, where $u\in\{a, v\}$. Then, $g^{*}(t,n)$ = $g_a(t,n) + g_v(t,n)$. As expected, with only $\mathcal{D}_{KL}^{a}$ and visual corruptions in Kinetics50-C, we see a drop in performance since retrieval is independent of the corrupted video modality. In general, there is a need to capture both audio and visual distances via $\mathcal{D}_{KL}^{a+v}$.

\begin{figure}[t!]
    \centering
    \begin{subfigure}[t]{0.48\columnwidth}
        \centering
        \includegraphics[width=\linewidth]{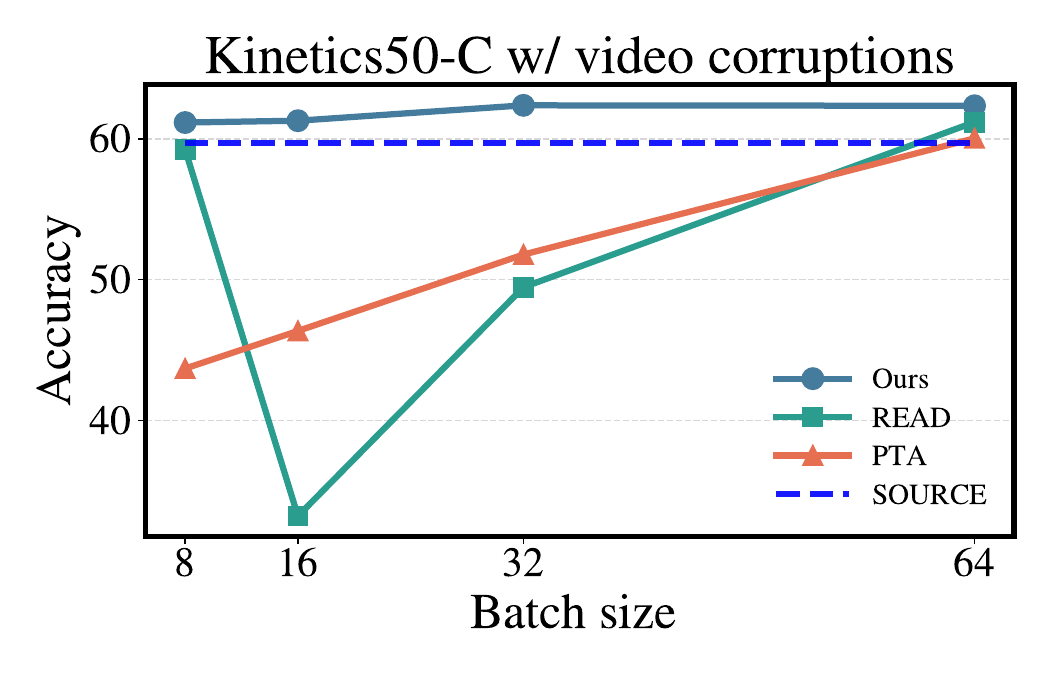}
        \caption{Kinetics50-C (video corruptions)}
        \label{fig:ks50-batch}
    \end{subfigure}
    \hfill
    \begin{subfigure}[t]{0.48\columnwidth}
        \centering
        \includegraphics[width=\linewidth]{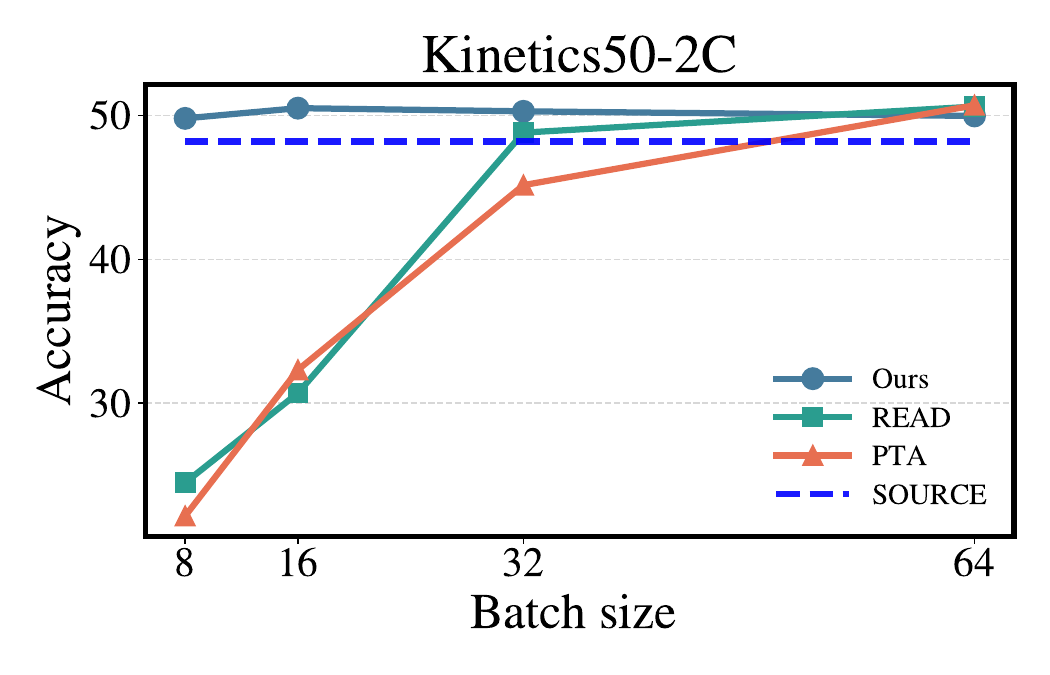}
        \caption{Kinetics50-2C (bimodal corruptions)}
        \label{fig:ks502c-batch}
    \end{subfigure}

    \caption{\textbf{$\framework$ maintains stable performance under small batch sizes.}
    The x-axis shows batch size, and the y-axis reports mean accuracy for $\framework$ and baselines.}
    \vspace{-10pt}
\end{figure}

\noindent \textbf{Effect of batch size.}\label{batch_size_ablation}
Here, we investigate the sensitivity of $\framework$ to the choice of batch size $B$. We evaluate a range of values $B \in \{8, 16, 32, 64\}$ to assess the framework’s robustness across varying computational constraints. The effect of batch size can be critical in our setting. In Eqns. \eqref{stats_v}, \eqref{stats_v_cov}, \eqref{stats_a}, and \eqref{stats_a_cov}, the mean and covariance are computed along the batch axis, where a smaller batch size can \textit{potentially} introduce noise into these statistics. We demonstrate results on Kinetics50-C (w/ video corruptions) and Kinetics50-2C. Our results in Figures \ref{fig:ks50-batch} and \ref{fig:ks502c-batch} demonstrate that $\framework$ maintains competitive performance even at $B=8$. This highlights the suitability of our proposed method under memory limitations, especially on small data streams. 

\noindent \textbf{Note.} In the Supplementary, we present additional experimental results and analyses.

\vspace{-8pt}
\section{Conclusion}
\label{conclusion}

In this paper, we comprehensively study audio-visual continual test-time adaptation, taking a major step forward in the real-time deployment of audio-visual models in continually changing target environments. We propose $\framework$, a novel method to address the same. Our motivation stems from the fact that the attention fusion layer exhibits intra- and cross-task transferability, suggesting that attention parameters from previous domains can be retrieved, plugged, and reused for continual adaptation. Inspired by the replay-based continual learning literature \cite{rebuffi2017icarl}, we introduce a shared buffer across tasks to store and selectively retrieve these generalizable attention parameters based on low-level input statistics. This effectively mitigates catastrophic forgetting and offers a scalable, memory-efficient framework. Through extensive evaluations on benchmark datasets involving unimodal and bimodal audio-visual corruptions, we establish SOTA baselines for robust audio-visual continual adaptation.

\bibliographystyle{splncs04}
\bibliography{main}

\clearpage
\begin{center}
    \Huge{\textbf{Supplementary}}
\end{center}
\rule[0pt]{\columnwidth}{1pt}

\section{Additional Details of $\framework$}\label{sec:linsys}

\subsection{KL Divergence Between Modality-Specific Gaussian Distributions}\label{proof}

In \S \ref{method:retrieval}, we utilize the KL divergence as a distance metric between the current modality-specific statistics and the stored buffer elements. We define the distributions at time-step $t$ and for the $n^{th}$ buffer element as $P_u^t = \mathcal{N}(\mu_u^t, \Sigma_u^t)$ and $P_u^n = \mathcal{N}(\mu_u^n, \Sigma_u^n)$, respectively. For simplicity and to ensure real-time retrieval efficiency as the buffer grows, we assume a diagonal covariance structure, $\Sigma = \text{diag}(\sigma_1^2, \cdots, \sigma_r^2)$. The PDF of a multivariate Gaussian is defined as $p(x) = (2\pi)^{-r/2} |\Sigma|^{-1/2} \exp\left( -\frac{1}{2} (x-\mu)^\top \Sigma^{-1} (x-\mu) \right)$, where $r$ is the feature dimension.
The closed-form solution for the KL divergence is derived as follows:

\begin{align}\label{dkl}
g_u(t, n) &= \mathcal{D}_{KL}(P_u^t \parallel P_u^n) = \mathbb{E}_{P_u^t} \left[ \frac{\ln p_u^t(x)}{\ln p_u^n(x)}\right]
\end{align} 
Now, upon substituting the PDF of a multivariate Gaussian in Eqn. \eqref{dkl}, 
\begin{align}
    \ln \frac{p_u^t(x)}{p_u^n(x)} = \frac{1}{2} \ln \frac{|\Sigma_u^n|}{|\Sigma_u^t|} - \frac{1}{2} (x-\mu_u^t)^\top (\Sigma_u^t)^{-1} (x-\mu_u^t) \\ + \frac{1}{2} (x-\mu_u^n)^\top (\Sigma_u^n)^{-1} (x-\mu_u^n)
\end{align}
This simplifies to, 
\begin{align}
    g_u(t, n) = \frac{1}{2} \ln \frac{|\Sigma_u^n|}{|\Sigma_u^t|} - \frac{1}{2} \mathbb{E}_{P_u^t}\left[(x-\mu_u^t)^\top (\Sigma_u^t)^{-1} (x-\mu_u^t)\right] + \\ \frac{1}{2} \mathbb{E}_{P_u^t}\left[(x-\mu_u^n)^\top (\Sigma_u^n)^{-1} (x-\mu_u^n)\right] \label{exp}
\end{align}

Using the identity $\mathbb{E}_{P}[x^\top A x] = \text{tr}(A\Sigma) + \mu^\top A \mu$ for a distribution with mean $\mu$ and covariance $\Sigma$, we evaluate the quadratic forms as, $\mathbb{E}_{P_u^t} [(x-\mu_u^t)^\top (\Sigma_u^t)^{-1} (x-\mu_u^t)] = \text{tr}((\Sigma_u^t)^{-1} \Sigma_u^t) = \text{tr}(\mathbf{I}_r) = r$. Similarly, $\mathbb{E}_{P_u^t} [(x-\mu_u^n)^\top (\Sigma_u^n)^{-1} (x-\mu_u^n)] = \text{tr}((\Sigma_u^n)^{-1} \Sigma_u^t) + (\mu_u^t - \mu_u^n)^\top (\Sigma_u^n)^{-1} (\mu_u^t - \mu_u^n)$. Upon substituting these into Eqn. \eqref{exp}, we obtain

\begin{equation}
\begin{aligned} \label{kl_gauss}
g_u(t, n) = \frac{1}{2} \left[ \underbrace{\text{tr}((\Sigma_u^n)^{-1} \Sigma_u^t)}_\text{\small{A}} + \underbrace{(\mu_u^t - \mu_u^n)^\top (\Sigma_u^n)^{-1} (\mu_u^t - \mu_u^n)}_\text{\small{B}} - r + \underbrace{\ln \frac{|\Sigma_u^n|}{|\Sigma_u^t|}}_\text{\small{C}} \right]
\end{aligned}    
\end{equation}

In our formulation, we assume a diagonal covariance matrix to maintain computational efficiency. Under this assumption, the individual terms simplify substantially. Term A in Eqn. \eqref{kl_gauss} reduces to $\sum_{i=1}^r \frac{(\sigma_{u,i}^t)^{2}}{(\sigma_{u,i}^n)^{2}}$, \ie a sum over diagonal entries only. Term B reduces to a weighted Euclidean distance as $\sum_{i=1}^r \frac{(\mu_{u,i}^t - \mu_{u,i}^n)^2}{(\sigma_{u,i}^n)^{2}}$. In addition, log-det term C simplifies as the log-product of diagonal elements as $\ln \frac{\prod (\sigma_{u,i}^n)^{2}}{\prod (\sigma_{u,i}^t)^{2}} = \sum_{i=1}^r \ln \frac{(\sigma_{u,i}^n)^{2}}{(\sigma_{u,i}^t)^{2}} = 2 \sum_{i=1}^r \ln \frac{\sigma_{u,i}^n}{\sigma_{u,i}^t}$. Putting all of them back in Eqn. \eqref{exp}, 

\begin{equation}
    \begin{aligned}
        g_u(t, n) = \frac{1}{2}\sum_{i=1}^{r}  \left[ \frac{(\sigma_{u,i}^t)^{2}}{(\sigma_{u,i}^n)^{2}} + \frac{(\mu_{u,i}^t - \mu_{u,i}^n)^2}{(\sigma_{u,i}^n)^{2}} -1 + \ln \frac{\sigma_{u,i}^n}{\sigma_{u,i}^t}\right] \\ 
         = \sum_{i=1}^{r} \left[\frac{(\sigma_{u,i}^t)^{2} + (\mu_{u,i}^t - \mu_{u,i}^n)^2}{2 (\sigma_{u,i}^n)^{2}} - \frac{1}{2} + \ln \frac{\sigma_{u,i}^n}{\sigma_{u,i}^t} \right]
    \end{aligned}
\end{equation}
For audio, $r=$ 128 and for visual frames, $r=$ 3.

\subsection{$\framework$ Algorithm}\label{alg:cap}
In Alg. \ref{working_algo}, we present the full pipeline of our proposed $\framework$ for audio-visual CTTA.

\begin{algorithm}[htb!]
    \caption{\framework: Selective Parameter Retrieval and Buffer Expansion}
    \begin{algorithmic}[1]
        \STATE  \textbf{Inputs:} Multimodal batch at time-step \textit{t}, $x^t=\{(x_a^{t,i}, x_v^{t,i})\}_{i=1}^B$, model parameters $\theta^{t-1}$ (where $\theta \in \{\mathcal{W}_Q, \mathcal{W}_K, \mathcal{W}_V\}$), threshold $\tau$, buffer size $\eta$, EMA smoothing factor $\beta$, learning rate $\alpha$
        \STATE \textbf{Outputs:} Predictions $y_t = \{y_i\}_{i=1}^B$
        \STATE Initialize empty buffer $\mathcal{K} \leftarrow \emptyset$
        \STATE Compute input-level statistics $(\mu_u^t, \Sigma_u^t)$ for $u \in \{a,v\}$ \hfill {\normalsize \color{deepcarminepink} \# Eqns. \eqref{stats_v}, \eqref{stats_v_cov}, \eqref{stats_a}, \eqref{stats_a_cov}}
        \STATE $n^* \leftarrow \varnothing$, $\mathrm{dist} \leftarrow \infty$
        \IF{$\mathcal{K}$ is empty}
            \STATE $\mathcal{K} \leftarrow \mathcal{K} \cup \{(\mu_u^t, \Sigma_u^t, \theta)\} $ for $u \in \{a,v\}$
            \STATE \textbf{continue}
        \ENDIF
        \IF{$\mathcal{K}$ is not empty}
         \STATE $n^*, \mathrm{dist} \leftarrow \arg\min_{n \in \mathcal{K}} (\mathcal{D}_{KL}^{a}\!\left(\mathcal{N}_a^t \parallel \mathcal{N}_a^n\right) + \mathcal{D}_{KL}^{v}\!\left(\mathcal{N}_v^t \parallel \mathcal{N}_v^n\right)) $
        \ENDIF
        \IF{$\mathrm{dist} < \tau$}
            \STATE $\theta \leftarrow \mathcal{K}[n^*].\mathrm{params}$  
            \quad \hfill {\normalsize \color{deepcarminepink} \# Selective Parameter Retrieval of $\mathcal{W}_Q, \mathcal{W}_K, \mathcal{W}_V$} 
            \STATE EMA update $\mathcal{K}[n^*].\mathrm{params}$ \hfill {\normalsize \color{deepcarminepink} \# Eqn. \eqref{param_ema}}
            \STATE EMA update $\mathcal{K}[n^*].\mathrm{stats}$ \hfill {\normalsize \color{deepcarminepink} \# Eqns. 
            \eqref{input_ema}, \eqref{input_ema1}}
        \ELSE
            \IF{$|\mathcal{K}| \ge \eta$}
                \STATE $\mathcal{K} \leftarrow \textsc{MergeClosestElements}(\mathcal{K})$
            \ENDIF
            \STATE $\mathcal{K} \leftarrow \mathcal{K} \cup \{(\mu_u^t, \Sigma_u^t, \theta)\} $ for $u \in \{a,v\}$
        \ENDIF
        \STATE $\mathcal{L} \leftarrow \mathcal{L}_{conf}\!\left(f_\theta(x_a^t,x_v^t)\right) - \mathcal{L}_{ne}\!\left(f_\theta(x_a^t,x_v^t)\right)$
        \STATE $\theta^{t} \leftarrow \theta^{t-1} - \alpha \nabla_\theta \mathcal{L}$  \quad \hfill {\normalsize \color{deepcarminepink} \#  Adaptation step of $\mathcal{W}_Q, \mathcal{W}_K, \mathcal{W}_V$}
        \STATE $y_t \leftarrow f_\theta(x_a^t,x_v^t)$
    \end{algorithmic}
    \label{working_algo}
\end{algorithm}

\section{Experimental Details}\label{add_exps}

\subsection{Datasets} \label{app_dataset} 
Our experiments are conducted on the test sets of popular audio-visual datasets like Kinetics50 \cite{carreira2017quo}, and VGGSound \cite{chen2020vggsound}, following READ \cite{yang2024test} and other baselines. Specifically, we adopt Kinetics50-C and VGGSound-C from READ, which contain unimodal corruptions and include 50 and 309 classes, respectively.
Kinetics50-C includes 2,466, while VGGSound-C contains 14,046 audio–visual test pairs. We apply 15 visual \cite{hendrycks2019benchmarking} and 6 audio corruptions at a severity level of 5, following READ. For the bimodal corruption setting, we borrow the 15 corruptions from AVRobustBench \cite{maharana2025texttt}, \ie \textit{Gaussian, Impulse, Shot, Speckle \dots Interference}, and apply them to the Kinetics50 and VGGSound source test sets. We name them Kinetics50-2C and VGGSound-2C, respectively, following their nomenclature. These reflect \textit{realistic}, \textit{co-occurring}, and \textit{correlated} bimodal corruptions, giving the feel of a real-world setting. \textit{It is worth mentioning that Kinetics50 \cite{carreira2017quo} and VGGSound \cite{chen2020vggsound} are visual-dominant and audio-dominant, respectively.} This means that task-specific information is in the visual and audio modalities, respectively.

\subsection{Source Model Architecture}
Following all prior AV-TTA methods, we use CAV-MAE \cite{gong2022contrastive} as the default audio-visual recognition model. CAV-MAE consists of 11-layer transformer encoders for both audio and visual modalities. Each video clip is sampled at 10 frames, with one randomly selected frame used as input to the visual encoder, while the corresponding 10-second audio waveform is converted into a spectrogram and processed by the audio encoder. The joint-encoder has only 1 transformer layer. Following prior work \cite{yang2024test}, spectrogram inputs are normalized using a dataset mean of $-$5.081 and a standard deviation of 4.4849.

\subsection{Implementation of the Baselines}\label{impl_base}
Here, we outline the implementation schemes of all the baselines used in our study. Throughout, we use a batch size of 32. 

\noindent \textbf{SOURCE\footnote{https://github.com/YuanGongND/cav-mae} \cite{gong2022contrastive}}:
We follow \cite{yang2024test} and adopt CAV-MAE \cite{gong2022contrastive} as the source model, which is used for inference. We use pre-trained weights of Kinetics50 and VGGSound, as released by \cite{yang2024test}.

\noindent \textbf{TENT\footnote{https://github.com/DequanWang/tent} \cite{wang2020tent}}:
We adapt all the affine parameters of all LayerNorms and minimize the entropy of model predictions with a learning rate of $1\times10^{-4}$ and an Adam optimizer.

\noindent \textbf{EATA\footnote{https://github.com/mr-eggplant/EATA} \cite{niu2022efficient}}:
With LayerNorm affine parameters being continually adapted, we optimize using Adam with a learning rate of $1\times10^{-4}$. The entropy threshold is set to 0.4$\times$log(C), where C denotes the number of classes. As source data is not accessible during adaptation, we do not apply Fisher regularization to preserve source-domain knowledge.

\noindent \textbf{SAR\footnote{https://github.com/mr-eggplant/SAR} \cite{niu2023towards}}: 
We continually adapt only the LayerNorm affine parameters, optimized with Adam at a constant learning rate of $1\times10^{-4}$. To stabilize entropy minimization, we employ the same confidence filtering strategy used in EATA \cite{niu2022efficient}, and trigger model recovery when the confidence drops below 0.2. Model predictions are smoothed using ENA with a momentum coefficient of 0.9.

\noindent \textbf{READ\footnote{https://github.com/XLearning-SCU/2024-ICLR-READ} \cite{yang2024test}}:
Following their core implementation, we continually adapt $\mathcal{W}_Q$, $\mathcal{W}_K$, and $\mathcal{W}_V$ of the joint-encoder with a learning rate of $1\times10^{-4}$ and an Adam optimizer. 

\noindent \textbf{SuMi\footnote{https://github.com/zrguo/SuMi} \cite{guo2025smoothing}}: We use learning rates of $1\times10^{-4}$ and $1\times10^{-5}$ on the derivatives of Kinetics50 and VGGSound, respectively, with an Adam optimizer. Throughout, LayerNorm affine parameters are continually adapted. For bimodal corruptions, the mutual information loss is applied once every two iterations. All remaining hyperparameters follow the original paper.

\noindent \textbf{PTA\footnote{https://github.com/MPI-Lab/PTA} \cite{wangpartition}}: For the test sets from Kinetics50, we use a recommended learning rate of $2\times10^{-4}$ and $1\times10^{-4}$ for VGGSound derivatives. $\mathcal{W}_Q$, $\mathcal{W}_K$, and $\mathcal{W}_V$ of the joint-encoder are continually adapted using the loss-specific hyperparameters recommended in the original method.

\noindent \textbf{BriMPR*\footnote{https://github.com/Luchicken/BriMPR} \cite{li2025bridging}}: As discussed earlier, we remove access to the 32 source-domain samples. We use 10 prompt tokens for both the audio and visual encoders. During adaptation, the visual prompts are continually adapted using a cross-modal masked embedding loss together with an instance-wise contrastive loss. The mask ratio is set to 0.5, and the contrastive temperature is fixed to 0.7. Additional implementation details follow the original paper.

\subsection{Implementation Details of $\framework$}\label{impl_ours}
At the onset of adaptation, we initialize a shared rehearsal buffer, $\mathcal{K} = \emptyset$. Each element in $\mathcal{K}$ is a tuple of modality-specific statistics and their corresponding adapted parameters: $(\mu_v^i, \Sigma_v^i, \mu_a^i, \Sigma_a^i, \mathcal{W}_Q^i, \mathcal{W}_K^i, \mathcal{W}_V^i)$. Here, $\mu$ and $\Sigma$ denote the mean and covariance for the visual ($v$) and audio ($a$) modalities, while $\{\mathcal{W}_Q, \mathcal{W}_K, \mathcal{W}_V\}$ are the Query, Key, and Value projection weights of the attention fusion layer. We use a batch size of 32 and optimize with a learning rate of $1\times10^{-4}$ using an Adam optimizer, for all the benchmark datasets. On datasets involving unimodal corruptions, \ie Kinetics50-C and VGGSound-C \cite{yang2024test}, we set $\tau$ to be 0.005. For Kinetics50-C and VGGSound-2C \cite{maharana2025texttt} that involve bimodal corruptions, $\tau$ is set to be 0.01.

\begin{table*}[ht!]
\centering
\setlength{\tabcolsep}{2.5pt}
\scriptsize
\caption{\textbf{$\framework$ achieves SOTA results on Kinetics50-C} with \textbf{video corruptions} (unimodal). We report the task-wise accuracy (\%) at a severity level of 5, in the \textbf{continual} setting. We also report results with varying buffer sizes $\eta$.}
\begin{adjustbox}{width=\textwidth}
\begin{tabular}{cccccccccccccccccc}
\toprule
\textbf{} & \textbf{Method} &
\multicolumn{3}{c}{\textbf{Noise}} &
\multicolumn{4}{c}{\textbf{Blur}} &
\multicolumn{4}{c}{\textbf{Weather}} &
\multicolumn{4}{c}{\textbf{Digital}} &
\textbf{Mean} \\
\cmidrule(lr){3-5} \cmidrule(lr){6-9} \cmidrule(lr){10-13} \cmidrule(lr){14-17}
\textbf{} & &
Gaussian & Shot & Impulse &
Defocus & Glass & Motion & Zoom &
Snow & Frost & Fog & Brightness &
Contrast & \shortstack{Elastic\\Transform} & Pixelate & JPEG &
\\
\midrule
\multirow{14}{*}{\rotatebox[origin=c]{90}{\normalsize Kinetics50-C}} &
SOURCE &
46.55 & 47.36 & 46.51 & 67.23 & 61.70 & 70.67 & 66.39 & 62.22 & 60.54 & 47.32 & 74.88 & 51.36 & 64.14 & 66.39 & 61.98 & 59.68
\\
& TENT &
45.55 & 42.51 & 27.84 &
22.76 & 8.33 & 3.64 & 2.24 &
2.40 & 2.08 & 2.00 & 2.04 &
2.28 & 2.04 & 2.00 & 2.00 &
11.31 \\
& EATA &
46.80 & 47.62 & 46.81 & 67.62 & 63.02 & 70.77 & 66.94 & 60.65 & 60.65 & 48.23 & 75.48 & 52.42 & 65.40 & 66.45 & 62.22 & 60.07
\\
& SAR &
46.83 & 47.00 & 46.64 & 65.71 & 62.30 & 69.99 & 66.35 & 60.54 & 60.22 & 52.60 & 73.96 & 50.76 & 65.99 & 63.42 & 59.90 & 59.48
\\
& READ &
49.96 & 51.80 & 51.56 & 68.07 & 65.63 & 65.38 & 61.50 & 52.20 & 49.48 & 42.19 & 54.93 & 28.93 & 39.82 & 33.17 & 27.32 & 49.46
\\
& SuMi &
45.75 & 45.43 & 44.95 & 64.30 & 65.38 & 69.23 & 67.91 & 62.86 & 66.67 & 63.22 & 69.23 & 50.16 & 71.15 & 66.31 & 63.94 & 61.10
\\
& PTA &
46.84 & 45.95 & 45.91 & 56.17 & 55.73 & 55.49 & 54.81 & 51.80 & 52.60 & 50.24 & 54.73 & 49.80 & 53.32 & 52.24 & 51.12 & 51.78
\\
& BriMPR* &
49.12 & 49.48 & 48.64 & 58.25 & 59.82 & 58.57 & 58.53 & 53.69 & 55.49 & 54.33 & 59.13 & 50.32 & 59.98 & 57.77 & 57.61 & 55.38
\\
& $\framework$ ($\eta=50$) &
\cellcolor[HTML]{C6F6D5}48.88 & \cellcolor[HTML]{C6F6D5}51.00 & \cellcolor[HTML]{C6F6D5}48.72 &
\cellcolor[HTML]{C6F6D5}68.23 & \cellcolor[HTML]{C6F6D5}64.78 & \cellcolor[HTML]{C6F6D5}70.99 & \cellcolor[HTML]{C6F6D5}68.47 &
\cellcolor[HTML]{C6F6D5}65.02 & \cellcolor[HTML]{C6F6D5}66.35 & \cellcolor[HTML]{C6F6D5}61.58 & \cellcolor[HTML]{C6F6D5}72.52 &
\cellcolor[HTML]{C6F6D5}53.81 & \cellcolor[HTML]{C6F6D5}66.79 & \cellcolor[HTML]{C6F6D5}68.19 & \cellcolor[HTML]{C6F6D5}65.99 &
\cellcolor[HTML]{C6F6D5}62.75
\\
& $\framework$ ($\eta=100$) &
\cellcolor[HTML]{C6F6D5}48.52 & \cellcolor[HTML]{C6F6D5}50.52 & \cellcolor[HTML]{C6F6D5}48.08 &
\cellcolor[HTML]{C6F6D5}67.95 & \cellcolor[HTML]{C6F6D5}63.86 & \cellcolor[HTML]{C6F6D5}71.59 & \cellcolor[HTML]{C6F6D5}67.75 &
\cellcolor[HTML]{C6F6D5}65.14 & \cellcolor[HTML]{C6F6D5}67.19 & \cellcolor[HTML]{C6F6D5}62.14 & \cellcolor[HTML]{C6F6D5}72.80 &
\cellcolor[HTML]{C6F6D5}54.17 & \cellcolor[HTML]{C6F6D5}66.43 & \cellcolor[HTML]{C6F6D5}67.59 & \cellcolor[HTML]{C6F6D5}65.18 &
\cellcolor[HTML]{C6F6D5}62.59
\\
& $\framework$ ($\eta=200$) &
\cellcolor[HTML]{C6F6D5}48.52 & \cellcolor[HTML]{C6F6D5}50.24 & \cellcolor[HTML]{C6F6D5}48.44 &
\cellcolor[HTML]{C6F6D5}67.91 & \cellcolor[HTML]{C6F6D5}63.18 & \cellcolor[HTML]{C6F6D5}71.68 & \cellcolor[HTML]{C6F6D5}68.03 &
\cellcolor[HTML]{C6F6D5}64.94 & \cellcolor[HTML]{C6F6D5}66.99 & \cellcolor[HTML]{C6F6D5}61.58 & \cellcolor[HTML]{C6F6D5}72.80 &
\cellcolor[HTML]{C6F6D5}53.97 & \cellcolor[HTML]{C6F6D5}66.11 & \cellcolor[HTML]{C6F6D5}67.87 & \cellcolor[HTML]{C6F6D5}64.42 &
\cellcolor[HTML]{C6F6D5}62.45 \\
& $\framework$ ($\eta=300$) &
\cellcolor[HTML]{C6F6D5}48.52 & \cellcolor[HTML]{C6F6D5}50.24 & \cellcolor[HTML]{C6F6D5}48.48 &
\cellcolor[HTML]{C6F6D5}67.91 & \cellcolor[HTML]{C6F6D5}63.18 & \cellcolor[HTML]{C6F6D5}71.60 & \cellcolor[HTML]{C6F6D5}68.07 &
\cellcolor[HTML]{C6F6D5}64.94 & \cellcolor[HTML]{C6F6D5}67.15 & \cellcolor[HTML]{C6F6D5}62.10 & \cellcolor[HTML]{C6F6D5}72.96 &
\cellcolor[HTML]{C6F6D5}53.57 & \cellcolor[HTML]{C6F6D5}65.67 & \cellcolor[HTML]{C6F6D5}67.87 & \cellcolor[HTML]{C6F6D5}64.54 &
\cellcolor[HTML]{C6F6D5}62.45 \\ 
& $\framework$ ($\eta=400$) &
\cellcolor[HTML]{C6F6D5}48.52 & \cellcolor[HTML]{C6F6D5}50.24 & \cellcolor[HTML]{C6F6D5}48.16 &
\cellcolor[HTML]{C6F6D5}67.91 & \cellcolor[HTML]{C6F6D5}63.22 & \cellcolor[HTML]{C6F6D5}71.60 & \cellcolor[HTML]{C6F6D5}67.99 &
\cellcolor[HTML]{C6F6D5}64.90 & \cellcolor[HTML]{C6F6D5}67.15 & \cellcolor[HTML]{C6F6D5}62.06 & \cellcolor[HTML]{C6F6D5}72.76 &
\cellcolor[HTML]{C6F6D5}53.57 & \cellcolor[HTML]{C6F6D5}65.55 & \cellcolor[HTML]{C6F6D5}67.83 & \cellcolor[HTML]{C6F6D5}64.66 &
\cellcolor[HTML]{C6F6D5}62.41 \\ 
& $\framework$ ($\eta=\infty$) &
\cellcolor[HTML]{C6F6D5}48.52 & \cellcolor[HTML]{C6F6D5}50.24 & \cellcolor[HTML]{C6F6D5}48.32 &
\cellcolor[HTML]{C6F6D5}67.91 & \cellcolor[HTML]{C6F6D5}63.18 & \cellcolor[HTML]{C6F6D5}71.60 & \cellcolor[HTML]{C6F6D5}68.03 &
\cellcolor[HTML]{C6F6D5}64.86 & \cellcolor[HTML]{C6F6D5}67.15 & \cellcolor[HTML]{C6F6D5}62.10 & \cellcolor[HTML]{C6F6D5}72.80 &
\cellcolor[HTML]{C6F6D5}53.57 & \cellcolor[HTML]{C6F6D5}65.43 & \cellcolor[HTML]{C6F6D5}67.87 & \cellcolor[HTML]{C6F6D5}64.02 &
\cellcolor[HTML]{C6F6D5}62.37
\\
\bottomrule

\multirow{14}{*}{\rotatebox[origin=c]{90}{\normalsize VGGSound-C}} &
SOURCE &
52.78 & 52.69 & 52.73 & 57.19 & 57.21 & 58.52 & 57.61 & 56.25 & 56.58 & 55.31 & 58.95 & 53.66 & 56.95 & 55.81 & 56.89 & 55.94
\\
& TENT &
52.58 & 51.82 & 51.20 & 53.85 & 53.90 & 54.49 & 54.39 & 51.80 & 52.39 & 52.59 & 52.14 & 51.08 & 52.49 & 51.70 & 51.87 & 52.55
\\
& EATA &
52.79 & 52.81 & 52.46 & 56.60 & 55.90 & 57.43 & 56.16 & 54.36 & 54.54 & 53.35 & 57.06 & 51.94 & 54.56 & 53.23 & 54.51 & 54.51
\\
& SAR &
52.92 & 52.86 & 52.92 & 57.03 & 56.66 & 58.65 & 57.61 & 55.95 & 56.74 & 56.22 & 58.25 & 54.21 & 57.13 & 55.42 & 56.46 & 55.94
\\
& READ &
52.43 & 52.65 & 52.44 & 55.85 & 55.22 & 56.07 & 55.69 & 54.51 & 54.96 & 54.69 & 55.38 & 54.34 & 54.79 & 54.65 & 54.65 & 54.55
\\
& SuMi & 
53.14 & 53.42 & 53.43 & 57.01 & 56.69 & 57.67 & 56.91 & 55.60 & 56.22 & 51.73 & 56.91 & 53.67 & 56.37 & 55.16 & 55.25 & 55.28\\
& PTA &
51.06 & 49.86 & 49.46 & 52.88 & 52.12 & 52.31 & 51.94 & 50.96 & 51.21 & 51.17 & 51.92 & 50.46 & 51.59 & 51.03 & 51.05 & 51.27
\\
& BriMPR* &
43.24 & 39.11 & 38.86 & 39.99 & 41.22 & 39.41 & 37.97 & 33.12 & 30.76 & 31.47 & 30.71 & 26.08 & 31.95 & 34.30 & 36.15 & 35.62
\\
& $\framework$ ($\eta=50$) &
\cellcolor[HTML]{C6F6D5}52.83 & \cellcolor[HTML]{C6F6D5}52.66 & \cellcolor[HTML]{C6F6D5}52.63 & \cellcolor[HTML]{C6F6D5}57.18 & \cellcolor[HTML]{C6F6D5}57.04 & \cellcolor[HTML]{C6F6D5}58.52 & \cellcolor[HTML]{C6F6D5}58.02 & \cellcolor[HTML]{C6F6D5}55.85 & \cellcolor[HTML]{C6F6D5}56.39 & \cellcolor[HTML]{C6F6D5}56.24 & \cellcolor[HTML]{C6F6D5}57.44 & \cellcolor[HTML]{C6F6D5}54.50 & \cellcolor[HTML]{C6F6D5}57.22 & \cellcolor[HTML]{C6F6D5}55.23 & \cellcolor[HTML]{C6F6D5}55.79 & \cellcolor[HTML]{C6F6D5}55.84 \\
& $\framework$ ($\eta=100$) &
\cellcolor[HTML]{C6F6D5}52.81 & \cellcolor[HTML]{C6F6D5}52.76 & \cellcolor[HTML]{C6F6D5}52.81 & \cellcolor[HTML]{C6F6D5}57.13 & \cellcolor[HTML]{C6F6D5}56.74 & \cellcolor[HTML]{C6F6D5}58.33 & \cellcolor[HTML]{C6F6D5}58.02 & \cellcolor[HTML]{C6F6D5}55.87 & \cellcolor[HTML]{C6F6D5}56.35 & \cellcolor[HTML]{C6F6D5}56.17 & \cellcolor[HTML]{C6F6D5}57.52 & \cellcolor[HTML]{C6F6D5}54.40 & \cellcolor[HTML]{C6F6D5}56.17 & \cellcolor[HTML]{C6F6D5}55.32 & \cellcolor[HTML]{C6F6D5}55.95 & \cellcolor[HTML]{C6F6D5}55.76
\\
& $\framework$ ($\eta=200$) &
\cellcolor[HTML]{C6F6D5}52.84 & \cellcolor[HTML]{C6F6D5}52.83 & \cellcolor[HTML]{C6F6D5}52.83 & \cellcolor[HTML]{C6F6D5}57.18 & \cellcolor[HTML]{C6F6D5}56.91 & \cellcolor[HTML]{C6F6D5}58.68 & \cellcolor[HTML]{C6F6D5}58.20 & \cellcolor[HTML]{C6F6D5}56.33 & \cellcolor[HTML]{C6F6D5}56.83 & \cellcolor[HTML]{C6F6D5}56.45 & \cellcolor[HTML]{C6F6D5}57.82 & \cellcolor[HTML]{C6F6D5}54.76 & \cellcolor[HTML]{C6F6D5}57.47 & \cellcolor[HTML]{C6F6D5}55.99 & \cellcolor[HTML]{C6F6D5}56.78  & \cellcolor[HTML]{C6F6D5}56.13\\
& $\framework$ ($\eta=300$) &
\cellcolor[HTML]{C6F6D5}52.87 & \cellcolor[HTML]{C6F6D5}52.87 & \cellcolor[HTML]{C6F6D5}52.86 & \cellcolor[HTML]{C6F6D5}57.16 & \cellcolor[HTML]{C6F6D5}56.97 & \cellcolor[HTML]{C6F6D5}58.65 & \cellcolor[HTML]{C6F6D5}58.02 & \cellcolor[HTML]{C6F6D5}56.66 & \cellcolor[HTML]{C6F6D5}57.13 & \cellcolor[HTML]{C6F6D5}56.67 & \cellcolor[HTML]{C6F6D5}58.18 & \cellcolor[HTML]{C6F6D5}54.66 & \cellcolor[HTML]{C6F6D5}57.42 & \cellcolor[HTML]{C6F6D5}56.22 & \cellcolor[HTML]{C6F6D5}56.71 & \cellcolor[HTML]{C6F6D5}56.20 \\
& $\framework$ ($\eta=400$) &
\cellcolor[HTML]{C6F6D5}52.86 & \cellcolor[HTML]{C6F6D5}52.83 & \cellcolor[HTML]{C6F6D5}52.91 & \cellcolor[HTML]{C6F6D5}57.30 & \cellcolor[HTML]{C6F6D5}57.25 & \cellcolor[HTML]{C6F6D5}58.66 & \cellcolor[HTML]{C6F6D5}57.95 & \cellcolor[HTML]{C6F6D5}56.50 & \cellcolor[HTML]{C6F6D5}56.76 & \cellcolor[HTML]{C6F6D5}56.47 & \cellcolor[HTML]{C6F6D5}57.89 & \cellcolor[HTML]{C6F6D5}54.80 & \cellcolor[HTML]{C6F6D5}57.40 & \cellcolor[HTML]{C6F6D5}55.82 & \cellcolor[HTML]{C6F6D5}56.54 & \cellcolor[HTML]{C6F6D5}56.13 \\
& $\framework$ ($\eta=\infty$) &
\cellcolor[HTML]{C6F6D5}52.87 & \cellcolor[HTML]{C6F6D5}52.83 & \cellcolor[HTML]{C6F6D5}52.91 &
\cellcolor[HTML]{C6F6D5}57.42 & \cellcolor[HTML]{C6F6D5}57.29 & \cellcolor[HTML]{C6F6D5}58.68 & \cellcolor[HTML]{C6F6D5}57.77 &
\cellcolor[HTML]{C6F6D5}56.38 & \cellcolor[HTML]{C6F6D5}56.80 & \cellcolor[HTML]{C6F6D5}55.50 & \cellcolor[HTML]{C6F6D5}59.13 &
\cellcolor[HTML]{C6F6D5}53.88 & \cellcolor[HTML]{C6F6D5}57.07 & \cellcolor[HTML]{C6F6D5}55.96 & \cellcolor[HTML]{C6F6D5}57.01 &
\cellcolor[HTML]{C6F6D5}56.10
\\
\bottomrule

\end{tabular}
\end{adjustbox}
\label{tab:budget_video}
\end{table*}

\begin{table*}[htb!]
\centering
\setlength{\tabcolsep}{2pt}
\scriptsize
\caption{\textbf{$\framework$ achieves SOTA results on Kinetics50-2C and VGGSound-2C} (bimodal). We report the task-wise accuracy (\%) at a severity level of 5, in the \textbf{continual} setting. We also report results with varying buffer sizes $\eta$.}
\begin{adjustbox}{width=\textwidth}
\begin{tabular}{cccccccccccccccccc}
\toprule
\textbf{} & \textbf{Method} &
\multicolumn{5}{c}{\textbf{Digital}} &
\multicolumn{6}{c}{\textbf{Environmental}} &
\multicolumn{4}{c}{\textbf{Human-Related}} &
\textbf{Mean} \\
\cmidrule(lr){3-7} \cmidrule(lr){8-13} \cmidrule(lr){14-17} 
\textbf{} & &
Gaussian & Impulse & Shot &
Speckle & Compression & Snow & Frost &
Spatter & Wind & Rain & Underwater &
Concert & Smoke & Crowd & Interference &
\\
\midrule
\multirow{14}{*}{\rotatebox[origin=c]{90}{\normalsize Kinetics50-2C}} &
SOURCE &
32.29 & 29.69 & 28.97 & 36.46 & 54.85 & 46.43 & 50.12 & 52.40 & 66.83 & 45.95 & 16.43 & 67.07 & 49.96 & 72.68 & 72.88 & 48.20
\\
& TENT &
22.76 & 3.93 & 2.00 & 2.12 & 6.81 & 1.96 & 2.00 & 2.00 & 2.36 & 1.96 & 1.96 & 1.96 & 1.96 & 2.00 & 1.96 & 3.85 \\
& EATA &
32.29 & 29.33 & 29.01 & 36.42 & 54.85 & 46.39 & 50.12 & 52.40 & 66.79 & 45.35 & 16.35 & 67.03 & 50.00 & 72.68 & 73.00 & 48.13
\\
& SAR &
32.05 & 29.53 & 28.12 & 35.54 & 53.40 & 45.67 & 49.12 & 50.44 & 64.50 & 44.15 & 17.43 & 64.38 & 47.32 & 69.99 & 72.20 & 46.92
\\
& READ &
37.62 & 36.74 & 38.22 & 42.83 & 59.66 & 48.64 & 55.25 & 52.48 & 59.02 & 47.52 & 37.86 & 57.33 & 40.87 & 58.53 & 59.58 & 48.81 
\\
& SuMi &
35.34 & 31.85 & 26.76 & 42.47 & 56.97 & 42.79 & 40.70 & 44.79 & 61.02 & 34.13 & 7.69 & 52.72 & 28.49 & 61.82 & 62.54 & 42.01
\\
& PTA &
33.13 & 30.81 & 31.37 & 31.97 & 46.59 & 40.70 & 46.87 & 45.27 & 56.29 & 45.91 & 39.18 & 58.41 & 44.71 & 62.78 & 63.46 & 45.16 
\\
& BriMPR* &
38.18 & 37.02 & 37.94 & 45.31 & 57.93 & 46.59 & 51.24 & 48.36 & 54.13 & 46.31 & 39.54 & 54.77 & 43.15 & 57.25 & 56.01 & 47.58
\\
&  $\framework$ ($\eta=50$) &
\cellcolor[HTML]{C6F6D5}34.13 & \cellcolor[HTML]{C6F6D5}30.77 & \cellcolor[HTML]{C6F6D5}33.73 &
\cellcolor[HTML]{C6F6D5}37.62 & \cellcolor[HTML]{C6F6D5}58.97 & \cellcolor[HTML]{C6F6D5}50.00 & \cellcolor[HTML]{C6F6D5}59.02 &
\cellcolor[HTML]{C6F6D5}57.01 & \cellcolor[HTML]{C6F6D5}65.38 & \cellcolor[HTML]{C6F6D5}56.17 & \cellcolor[HTML]{C6F6D5}19.35 &
\cellcolor[HTML]{C6F6D5}66.79 & \cellcolor[HTML]{C6F6D5}51.16 & \cellcolor[HTML]{C6F6D5}70.39 & \cellcolor[HTML]{C6F6D5}72.32 & \cellcolor[HTML]{C6F6D5}50.85 \\ 
&  $\framework$ ($\eta=100$) &
\cellcolor[HTML]{C6F6D5}34.13 & \cellcolor[HTML]{C6F6D5}31.01 & \cellcolor[HTML]{C6F6D5}33.69 &
\cellcolor[HTML]{C6F6D5}36.94 & \cellcolor[HTML]{C6F6D5}59.05 & \cellcolor[HTML]{C6F6D5}49.36 & \cellcolor[HTML]{C6F6D5}57.25 &
\cellcolor[HTML]{C6F6D5}56.85 & \cellcolor[HTML]{C6F6D5}66.07 & \cellcolor[HTML]{C6F6D5}56.21 & \cellcolor[HTML]{C6F6D5}13.30 &
\cellcolor[HTML]{C6F6D5}67.15 & \cellcolor[HTML]{C6F6D5}51.08 & \cellcolor[HTML]{C6F6D5}72.07 & \cellcolor[HTML]{C6F6D5}73.20 & \cellcolor[HTML]{C6F6D5}50.49 \\ 
&  $\framework$ ($\eta=200$) &
\cellcolor[HTML]{C6F6D5}34.13 & \cellcolor[HTML]{C6F6D5}30.73 & \cellcolor[HTML]{C6F6D5}33.69 &
\cellcolor[HTML]{C6F6D5}36.86 & \cellcolor[HTML]{C6F6D5}59.02 & \cellcolor[HTML]{C6F6D5}48.36 & \cellcolor[HTML]{C6F6D5}55.13 &
\cellcolor[HTML]{C6F6D5}55.73 & \cellcolor[HTML]{C6F6D5}66.83 & \cellcolor[HTML]{C6F6D5}55.61 & \cellcolor[HTML]{C6F6D5}13.34 &
\cellcolor[HTML]{C6F6D5}67.23 & \cellcolor[HTML]{C6F6D5}51.44 & \cellcolor[HTML]{C6F6D5}72.36 & \cellcolor[HTML]{C6F6D5}73.28 & \cellcolor[HTML]{C6F6D5}50.25 \\
&  $\framework$ ($\eta=300$) &
\cellcolor[HTML]{C6F6D5}34.13 & \cellcolor[HTML]{C6F6D5}31.01 & \cellcolor[HTML]{C6F6D5}33.77 &
\cellcolor[HTML]{C6F6D5}36.86 & \cellcolor[HTML]{C6F6D5}59.02 & \cellcolor[HTML]{C6F6D5}48.32 & \cellcolor[HTML]{C6F6D5}55.09 &
\cellcolor[HTML]{C6F6D5}56.29 & \cellcolor[HTML]{C6F6D5}66.95 & \cellcolor[HTML]{C6F6D5}55.21 & \cellcolor[HTML]{C6F6D5}12.74 &
\cellcolor[HTML]{C6F6D5}67.11 & \cellcolor[HTML]{C6F6D5}51.88 & \cellcolor[HTML]{C6F6D5}72.40 & \cellcolor[HTML]{C6F6D5}73.04 & \cellcolor[HTML]{C6F6D5}50.25 \\
&  $\framework$ ($\eta=400$) &
\cellcolor[HTML]{C6F6D5}34.13 & \cellcolor[HTML]{C6F6D5}30.81 & \cellcolor[HTML]{C6F6D5}33.77 &
\cellcolor[HTML]{C6F6D5}36.86 & \cellcolor[HTML]{C6F6D5}59.02 & \cellcolor[HTML]{C6F6D5}48.32 & \cellcolor[HTML]{C6F6D5}55.13 &
\cellcolor[HTML]{C6F6D5}56.61 & \cellcolor[HTML]{C6F6D5}66.87 & \cellcolor[HTML]{C6F6D5}55.09 & \cellcolor[HTML]{C6F6D5}13.90 &
\cellcolor[HTML]{C6F6D5}67.15 & \cellcolor[HTML]{C6F6D5}51.60 & \cellcolor[HTML]{C6F6D5}72.44 & \cellcolor[HTML]{C6F6D5}73.32 &
\cellcolor[HTML]{C6F6D5}50.33 \\
& $\framework$ ($\eta=\infty$) &
\cellcolor[HTML]{C6F6D5}34.13 & \cellcolor[HTML]{C6F6D5}31.09 & \cellcolor[HTML]{C6F6D5}33.73 &
\cellcolor[HTML]{C6F6D5}36.82 & \cellcolor[HTML]{C6F6D5}59.02 & \cellcolor[HTML]{C6F6D5}48.32 & \cellcolor[HTML]{C6F6D5}55.21 &
\cellcolor[HTML]{C6F6D5}56.57 & \cellcolor[HTML]{C6F6D5}66.87 & \cellcolor[HTML]{C6F6D5}55.13 & \cellcolor[HTML]{C6F6D5}14.50 &
\cellcolor[HTML]{C6F6D5}66.83 & \cellcolor[HTML]{C6F6D5}51.12 & \cellcolor[HTML]{C6F6D5}71.87 & \cellcolor[HTML]{C6F6D5}73.12 &
\cellcolor[HTML]{C6F6D5}50.29
\\
\bottomrule

\multirow{14}{*}{\rotatebox[origin=c]{90}{\normalsize VGGSound-2C}} &
SOURCE &
30.11 & 23.63 & 20.77 & 25.77 & 34.57 & 24.48 & 46.39 & 48.43 & 50.40 & 29.50 & 42.15 & 47.60 & 31.83 & 47.64 & 55.21 & 37.23
\\
& TENT &
2.87 & 0.33 & 0.33 & 0.37 & 0.89 & 0.33 & 0.33 & 0.33 & 0.33 & 0.33 & 0.33 & 0.33 & 0.33 & 0.33 & 0.33 & 0.54
\\
& EATA &
30.16 & 23.81 & 20.97 & 26.36 & 34.73 & 24.55 & 46.69 & 47.94 & 49.50 & 28.26 & 40.90 & 46.00 & 29.27 & 45.95 & 53.46 & 36.57
\\
& SAR &
12.74 & 10.90 & 13.30 & 18.97 & 36.62 & 18.21 & 43.16 & 46.75 & 49.01 & 19.70 & 41.63 & 31.10 & 23.26 & 43.85 & 51.09 & 30.69
\\
& READ &
43.74 & 38.89 & 40.23 & 26.72 & 25.54 & 22.65 & 27.92 & 29.26 & 30.81 & 22.08 & 25.53 & 26.18 & 25.04 & 28.38 & 33.09 & 29.74
\\
& SuMi & 
33.23 & 25.37 & 24.70 & 27.67 & 35.44 & 20.79 & 46.54 & 48.04 & 49.66 & 24.99 & 38.67 & 47.76 & 31.93 & 48.43 & 55.42 & 37.24
\\
& PTA &
40.82 & 34.35 & 33.69 & 28.21 & 29.71 & 32.52 & 39.27 & 41.44 & 41.59 & 32.35 & 36.44 & 36.41 & 34.52 & 35.96 & 40.37 & 35.84
\\
& BriMPR* &
30.47 & 24.53 & 25.02 & 24.12 & 26.15 & 21.54 & 27.77 & 26.88 & 26.73 & 20.81 & 23.24 & 23.55 & 18.83 & 25.85 & 28.20 & 24.91
\\
&  $\framework$ ($\eta=50$) &
 \cellcolor[HTML]{C6F6D5}40.63 & \cellcolor[HTML]{C6F6D5}34.77 & \cellcolor[HTML]{C6F6D5}37.70 &
\cellcolor[HTML]{C6F6D5}31.37 & \cellcolor[HTML]{C6F6D5}34.95 & \cellcolor[HTML]{C6F6D5}41.07 &
\cellcolor[HTML]{C6F6D5}50.02 & \cellcolor[HTML]{C6F6D5}49.03 & \cellcolor[HTML]{C6F6D5}52.51 &
\cellcolor[HTML]{C6F6D5}38.79 & \cellcolor[HTML]{C6F6D5}29.37 & \cellcolor[HTML]{C6F6D5}29.85 &
\cellcolor[HTML]{C6F6D5}24.06 & \cellcolor[HTML]{C6F6D5}28.40 & \cellcolor[HTML]{C6F6D5}35.58 &
\cellcolor[HTML]{C6F6D5}37.21 \\ 
&  $\framework$ ($\eta=100$) &
\cellcolor[HTML]{C6F6D5}40.42 & \cellcolor[HTML]{C6F6D5}33.95 & \cellcolor[HTML]{C6F6D5}37.66 &
\cellcolor[HTML]{C6F6D5}30.05 & \cellcolor[HTML]{C6F6D5}34.80 & \cellcolor[HTML]{C6F6D5}40.81 &
\cellcolor[HTML]{C6F6D5}50.07 & \cellcolor[HTML]{C6F6D5}49.86 & \cellcolor[HTML]{C6F6D5}52.58 &
\cellcolor[HTML]{C6F6D5}44.52 & \cellcolor[HTML]{C6F6D5}39.74 & \cellcolor[HTML]{C6F6D5}34.44 &
\cellcolor[HTML]{C6F6D5}28.85 & \cellcolor[HTML]{C6F6D5}31.85 & \cellcolor[HTML]{C6F6D5}37.68 & \cellcolor[HTML]{C6F6D5}39.15 \\ 
&  $\framework$ ($\eta=200$) &
\cellcolor[HTML]{C6F6D5}40.42 & \cellcolor[HTML]{C6F6D5}33.58 & \cellcolor[HTML]{C6F6D5}37.02 &
\cellcolor[HTML]{C6F6D5}29.60 & \cellcolor[HTML]{C6F6D5}34.63 & \cellcolor[HTML]{C6F6D5}40.33 &
\cellcolor[HTML]{C6F6D5}49.81 & \cellcolor[HTML]{C6F6D5}49.49 & \cellcolor[HTML]{C6F6D5}51.96 &
\cellcolor[HTML]{C6F6D5}44.63 & \cellcolor[HTML]{C6F6D5}45.90 & \cellcolor[HTML]{C6F6D5}49.47 &
\cellcolor[HTML]{C6F6D5}46.10 & \cellcolor[HTML]{C6F6D5}51.00 & \cellcolor[HTML]{C6F6D5}54.55 &
\cellcolor[HTML]{C6F6D5}43.90 \\
&  $\framework$ ($\eta=300$) &
\cellcolor[HTML]{C6F6D5}40.42 & \cellcolor[HTML]{C6F6D5}33.90 & \cellcolor[HTML]{C6F6D5}36.86 &
\cellcolor[HTML]{C6F6D5}29.57 & \cellcolor[HTML]{C6F6D5}34.61 & \cellcolor[HTML]{C6F6D5}40.22 &
\cellcolor[HTML]{C6F6D5}49.84 & \cellcolor[HTML]{C6F6D5}49.35 & \cellcolor[HTML]{C6F6D5}51.54 &
\cellcolor[HTML]{C6F6D5}44.31 & \cellcolor[HTML]{C6F6D5}45.71 & \cellcolor[HTML]{C6F6D5}49.34 &
\cellcolor[HTML]{C6F6D5}46.04 & \cellcolor[HTML]{C6F6D5}50.69 & \cellcolor[HTML]{C6F6D5}54.79 &
\cellcolor[HTML]{C6F6D5}43.81 \\
&  $\framework$ ($\eta=400$) &
\cellcolor[HTML]{C6F6D5}40.42 & \cellcolor[HTML]{C6F6D5}33.75 & \cellcolor[HTML]{C6F6D5}37.03 &
\cellcolor[HTML]{C6F6D5}29.55 & \cellcolor[HTML]{C6F6D5}34.59 & \cellcolor[HTML]{C6F6D5}40.07 &
\cellcolor[HTML]{C6F6D5}49.83 & \cellcolor[HTML]{C6F6D5}49.26 & \cellcolor[HTML]{C6F6D5}51.45 &
\cellcolor[HTML]{C6F6D5}44.29 & \cellcolor[HTML]{C6F6D5}45.65 & \cellcolor[HTML]{C6F6D5}49.20 &
\cellcolor[HTML]{C6F6D5}46.04 & \cellcolor[HTML]{C6F6D5}50.18 & \cellcolor[HTML]{C6F6D5}54.89 &
\cellcolor[HTML]{C6F6D5}43.75 \\
& $\framework$ ($\eta=\infty$) &
\cellcolor[HTML]{C6F6D5}40.42 & \cellcolor[HTML]{C6F6D5}33.78 & \cellcolor[HTML]{C6F6D5}36.79 &
\cellcolor[HTML]{C6F6D5}29.33 & \cellcolor[HTML]{C6F6D5}34.62 & \cellcolor[HTML]{C6F6D5}39.96 & \cellcolor[HTML]{C6F6D5}49.73 &
\cellcolor[HTML]{C6F6D5}49.12 & \cellcolor[HTML]{C6F6D5}51.12 & \cellcolor[HTML]{C6F6D5}44.21 & \cellcolor[HTML]{C6F6D5}45.34 &
\cellcolor[HTML]{C6F6D5}48.66 & \cellcolor[HTML]{C6F6D5}46.14 & \cellcolor[HTML]{C6F6D5}48.53 & \cellcolor[HTML]{C6F6D5}54.91 &
\cellcolor[HTML]{C6F6D5}43.51
\\
\bottomrule
\end{tabular}
\end{adjustbox}
\label{tab:bimodal_budget}
\end{table*}

\begin{table}[htb!]
\centering
\setlength{\tabcolsep}{2.5pt}
\scriptsize
\caption{\textbf{$\framework$ achieves SOTA results on Kinetics-50C and VGGSound-C} with \textbf{audio corruptions} (unimodal). We report the task-wise accuracy (\%) at a severity level of 5, in the \textbf{continual} setting. We also report results with varying buffer sizes $\eta$.}
\begin{adjustbox}{width=0.5\textwidth}
\begin{tabular}{ccccccccc}
\toprule
\textbf{} & \textbf{Method} &
\multicolumn{3}{c}{\textbf{Noise}} &
\multicolumn{3}{c}{\textbf{Weather}} &
\textbf{Mean} \\
\cmidrule(lr){3-5} \cmidrule(lr){6-8} 
\textbf{} & &
Gaussian & Traffic & Crowd &
Rain & Thunder & Wind &
\\
\midrule
\multirow{14}{*}{\rotatebox[origin=c]{90}{\normalsize Kinetics50-C}} &
SOURCE &
73.48 & 65.30 & 67.59 & 70.11 & 67.67 & 70.11 & 69.04 
\\
& TENT &
73.84 & 68.55 & 70.51 & 69.59 & 73.12 & 70.15 & 70.96 \\
& EATA &
73.56 & 65.30 & 67.71 & 70.11 & 68.07 & 70.15 & 69.15
\\
& SAR &
73.36 & 65.95 & 68.11 & 69.71 & 69.15 & 69.71 & 69.33
\\
& READ &
74.28 & 69.63 & 70.63 & 70.35 & 71.76 & 69.39 & 71.01
\\
& SuMi &
73.76 & 68.19 & 70.59 & 69.27 & 73.24 & 69.43 & 70.75
\\
& PTA &
72.68 & 69.03 & 69.79 & 68.83 & 71.79 & 69.75 & 70.31
\\
& BriMPR* &
72.84 & 67.67 & 67.43 & 65.95 & 69.07 & 63.70 & 67.78
\\
& $\framework$ ($\eta=50$) &\
\cellcolor[HTML]{C6F6D5}73.72 & \cellcolor[HTML]{C6F6D5}68.35 & \cellcolor[HTML]{C6F6D5}70.27 &
\cellcolor[HTML]{C6F6D5}70.11 & \cellcolor[HTML]{C6F6D5}72.88 & \cellcolor[HTML]{C6F6D5}69.91 &
\cellcolor[HTML]{C6F6D5}70.87 \\
& $\framework$ ($\eta=100$) &\
\cellcolor[HTML]{C6F6D5}73.72 & \cellcolor[HTML]{C6F6D5}68.47 & \cellcolor[HTML]{C6F6D5}69.79 &
\cellcolor[HTML]{C6F6D5}70.27 & \cellcolor[HTML]{C6F6D5}72.56 & \cellcolor[HTML]{C6F6D5}70.39 &
\cellcolor[HTML]{C6F6D5}70.87  \\
& $\framework$ ($\eta=200$) &\
\cellcolor[HTML]{C6F6D5}73.72 & \cellcolor[HTML]{C6F6D5}68.47 & \cellcolor[HTML]{C6F6D5}69.87 &
\cellcolor[HTML]{C6F6D5}70.19 & \cellcolor[HTML]{C6F6D5}71.63 & \cellcolor[HTML]{C6F6D5}70.51 &
\cellcolor[HTML]{C6F6D5}70.73 \\
& $\framework$ ($\eta=300$) &\
\cellcolor[HTML]{C6F6D5}73.72 & \cellcolor[HTML]{C6F6D5}68.47 & \cellcolor[HTML]{C6F6D5}69.87 &
\cellcolor[HTML]{C6F6D5}70.19 & \cellcolor[HTML]{C6F6D5}71.63 & \cellcolor[HTML]{C6F6D5}70.15 &
\cellcolor[HTML]{C6F6D5}70.67 \\
& $\framework$ ($\eta=400$) &\
\cellcolor[HTML]{C6F6D5}73.72 & \cellcolor[HTML]{C6F6D5}68.47 & \cellcolor[HTML]{C6F6D5}69.87 &
\cellcolor[HTML]{C6F6D5}70.19 & \cellcolor[HTML]{C6F6D5}71.63 & \cellcolor[HTML]{C6F6D5}70.15 &
\cellcolor[HTML]{C6F6D5}70.67  \\
& $\framework$ ($\eta=\infty$) &
\cellcolor[HTML]{C6F6D5}74.15 & \cellcolor[HTML]{C6F6D5}68.90 & \cellcolor[HTML]{C6F6D5}70.30 &
\cellcolor[HTML]{C6F6D5}70.62 & \cellcolor[HTML]{C6F6D5}72.06 & \cellcolor[HTML]{C6F6D5}70.57 &
\cellcolor[HTML]{C6F6D5}71.10
\\
\midrule
\multirow{13}{*}{\rotatebox[origin=c]{90}{\normalsize  VGGSound-C}} &
SOURCE &
37.36 & 21.12 & 16.80 & 21.64 & 27.29 & 25.54 & 24.96
\\
& TENT &
6.20 & 0.46 & 0.29 & 0.28 & 0.28 & 0.28 & 1.30
\\
& EATA &
37.51 & 21.64 & 17.58 & 22.82 & 27.98 & 25.34 & 25.48
\\
& SAR &
36.53 & 8.01 & 4.12 & 4.49 & 13.43 & 3.34 & 11.65
\\
& READ &
28.01 & 15.10 & 17.35 & 13.69 & 20.24 & 14.37 & 18.13
\\
& SuMi &
37.66 & 19.28 & 14.79 & 20.73 & 28.82 & 28.05 & 24.89
\\
& PTA &
36.30 & 28.79 & 28.42 & 25.35 & 30.60 & 26.09 & 29.26
\\
& BriMPR* &
23.12 & 14.95 & 15.15 & 13.85 & 20.08 & 15.24 & 17.07
\\
& $\framework$ ($\eta=50$) &
\cellcolor[HTML]{C6F6D5}39.67 & \cellcolor[HTML]{C6F6D5}27.99 & \cellcolor[HTML]{C6F6D5}17.31 &
\cellcolor[HTML]{C6F6D5}16.99 & \cellcolor[HTML]{C6F6D5}19.05 & \cellcolor[HTML]{C6F6D5}14.12 &
\cellcolor[HTML]{C6F6D5}22.52 \\
& $\framework$ ($\eta=100$) &
\cellcolor[HTML]{C6F6D5}39.55 & \cellcolor[HTML]{C6F6D5}28.63 & \cellcolor[HTML]{C6F6D5}23.15 &
\cellcolor[HTML]{C6F6D5}25.50 & \cellcolor[HTML]{C6F6D5}21.49 & \cellcolor[HTML]{C6F6D5}15.34 &
\cellcolor[HTML]{C6F6D5}25.61\\
& $\framework$ ($\eta=200$) &
\cellcolor[HTML]{C6F6D5}39.55 & \cellcolor[HTML]{C6F6D5}28.42 & \cellcolor[HTML]{C6F6D5}24.45 &
\cellcolor[HTML]{C6F6D5}30.14 & \cellcolor[HTML]{C6F6D5}30.90 & \cellcolor[HTML]{C6F6D5}17.81 &
\cellcolor[HTML]{C6F6D5}28.54 \\
& $\framework$ ($\eta=300$) &
\cellcolor[HTML]{C6F6D5}39.52 & \cellcolor[HTML]{C6F6D5}28.43 & \cellcolor[HTML]{C6F6D5}24.36 &
\cellcolor[HTML]{C6F6D5}29.81 & \cellcolor[HTML]{C6F6D5}33.58 & \cellcolor[HTML]{C6F6D5}20.85 &
\cellcolor[HTML]{C6F6D5}29.42 \\
& $\framework$ ($\eta=400$) &
\cellcolor[HTML]{C6F6D5}39.52 & \cellcolor[HTML]{C6F6D5}28.43 & \cellcolor[HTML]{C6F6D5}24.32 &
\cellcolor[HTML]{C6F6D5}29.37 & \cellcolor[HTML]{C6F6D5}35.65 & \cellcolor[HTML]{C6F6D5}26.38 &
\cellcolor[HTML]{C6F6D5}30.61\\
& $\framework$ ($\eta=\infty$) &
\cellcolor[HTML]{C6F6D5}39.52 & \cellcolor[HTML]{C6F6D5}28.43 & \cellcolor[HTML]{C6F6D5}24.31 &
\cellcolor[HTML]{C6F6D5}29.44 & \cellcolor[HTML]{C6F6D5}35.52 & \cellcolor[HTML]{C6F6D5}27.05 &
\cellcolor[HTML]{C6F6D5}30.71
\\
\bottomrule
\end{tabular}
\end{adjustbox}
\label{tab:budget_audio}
\end{table}

\newpage
\section{Additional Results}
\subsection{Results on a fixed buffer budget $\eta$}\label{budget_study}

In Tables \ref{tab:budget_video} and \ref{tab:budget_audio}, we report additional results on Kinetics50-C and VGGSound, respectively. In Kinetics50-C, the video is corrupted, whereas in VGGSound, the audio is corrupted. As a reminder, the task-specific information lies in the visual and audio modalities for Kinetics50 and VGGSound, respectively. The results indicate that $\framework$ maintains superior performance even when the "primary" modality is corrupted.

We observe distinct scaling behaviors across these datasets. On Kinetics50-C, $\framework$ achieves stable and competitive performance with a buffer size as small as $\eta=50$. Given that the test set comprises $36,990$ samples ($2,466 \times 15$ tasks), a budget of $\eta=50$ represents a compression ratio of nearly $740:1$, yet it is sufficient. Conversely, on VGGSound-C, comparable or superior performance is achieved at $\eta \ge 200$. We attribute this to the size. VGGSound-C has a significantly larger amount of data ($14046\times6$ tasks), necessitating a slightly larger set of buffer elements to adequately cover the expanded distribution. Despite these differences in scale, the overall results confirm that $\framework$ effectively bounds memory growth without sacrificing adaptation quality, making it ideal for large-scale, long-term deployment.

\subsection{Task-wise results of bimodal corruptions }\label{addn_task}

In Table \ref{tab:bimodal_budget}, we report the task-wise accuracies on Kinetics50-2C (top) and VGGSound-2C (bottom). For our proposed $\framework$, we also report the performances for various buffer sizes. Despite the increased complexity of the bimodal shifts, $\framework$ consistently achieves superior performance compared to existing baselines. Our rationale and findings remain the same as in \S \ref{budget_study}. Over a wide range of buffer budgets, we observe consistent performances on Kinetics50-2C, which has about $2,466 \times 15$ test samples. On a larger dataset like VGGSound-2C, $\eta\ge100$ achieves SOTA results.

\subsection{Alternatives to maintaining a buffer budget}\label{budget_alt}
In our proposed method, to maintain a strict budget $\eta$ of buffer $\mathcal{K}$, we perform statistical merging of the most redundant elements. We perform pairwise comparisons of all stored elements in $\mathcal{K}$ by solving Eqn. \eqref{merge_eq}. Here, we explore another strategy following \cite{zhang2024dpcore}, where the oldest element in $\mathcal{K}$ is removed when $|\mathcal{K}|\ge\eta$.  We conduct experiments on Kinetics50-2C and VGGSound-2C. In Table \ref{tab:merge_alt}, we observe that removing the oldest element maintains robust performance for both datasets involving bimodal corruptions. On VGGSound-2C, with $\eta$ as low as 50, $\framework$ maintains a stronger performance compared to merging by averaging elements (see Table \ref{tab:bimodal_budget}). However, we do posit that this strategy is highly dependent on the order of tasks. In settings where tasks arrive in a random order, discarding the oldest element may prove detrimental. This risks prematurely removing statistically relevant elements that belong to corruption categories, which may recur later in the stream, as evidenced by the cross-task transferability observed in \S \ref{analysis}.

\begin{table}[t!]
\centering
\setlength{\tabcolsep}{2pt}
\small
\caption{\textbf{$\framework$ remains stable in performance when the oldest element is removed from the buffer to maintain the budget.} We report the mean accuracies on Kinetics50-2C and VGGSound-2C with varying buffer sizes $\eta$.}
\begin{adjustbox}{width=0.5\textwidth}
\begin{tabular}{ccc}
\toprule
\textbf{Buffer budget $\eta$} &
\textbf{Kinetics50-2C} & \textbf{VGGSound-2C}  \\
\midrule
$\eta=50$ & 50.38 & 44.10 \\
$\eta=100$ & 50.43 & 43.81 \\
$\eta=200$ & 50.28 & 43.51  \\
\bottomrule
\end{tabular}
\end{adjustbox}
\label{tab:merge_alt}
\end{table}

\begin{figure*}[ht!]
    \centering
    \begin{subfigure}[t]{0.33\textwidth}
        \centering
        \includegraphics[width=\linewidth,trim={7mm 5mm 5mm 5mm},clip]{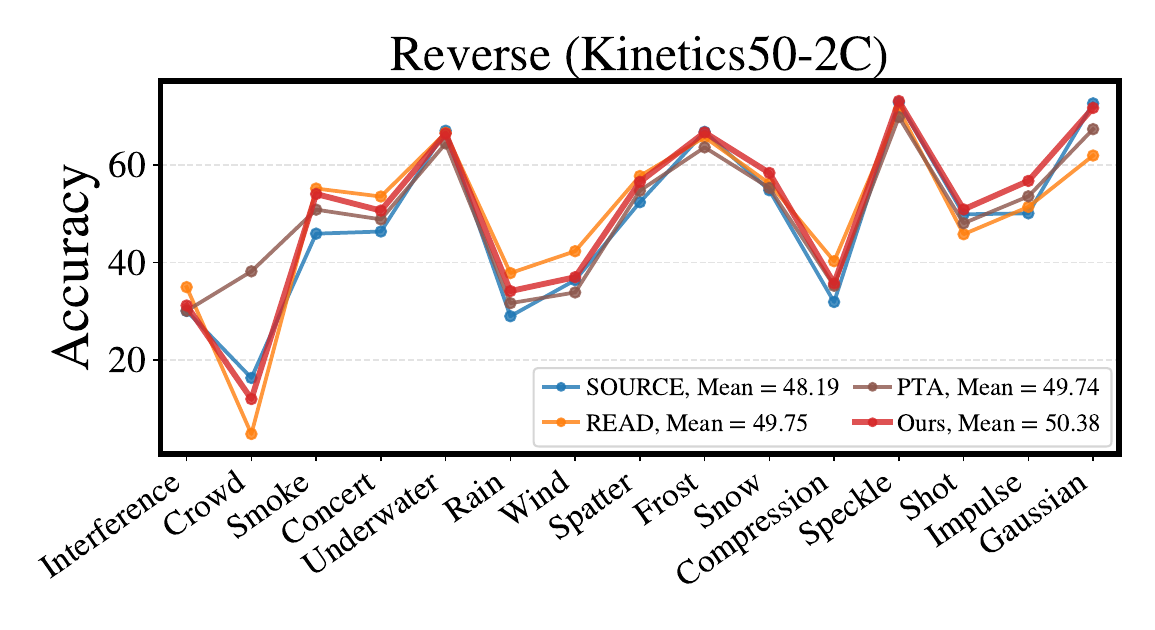}
    \end{subfigure}\hfill
    \begin{subfigure}[t]{0.33\textwidth}
        \centering
        \includegraphics[width=\linewidth,trim={7mm 5mm 5mm 5mm},clip]{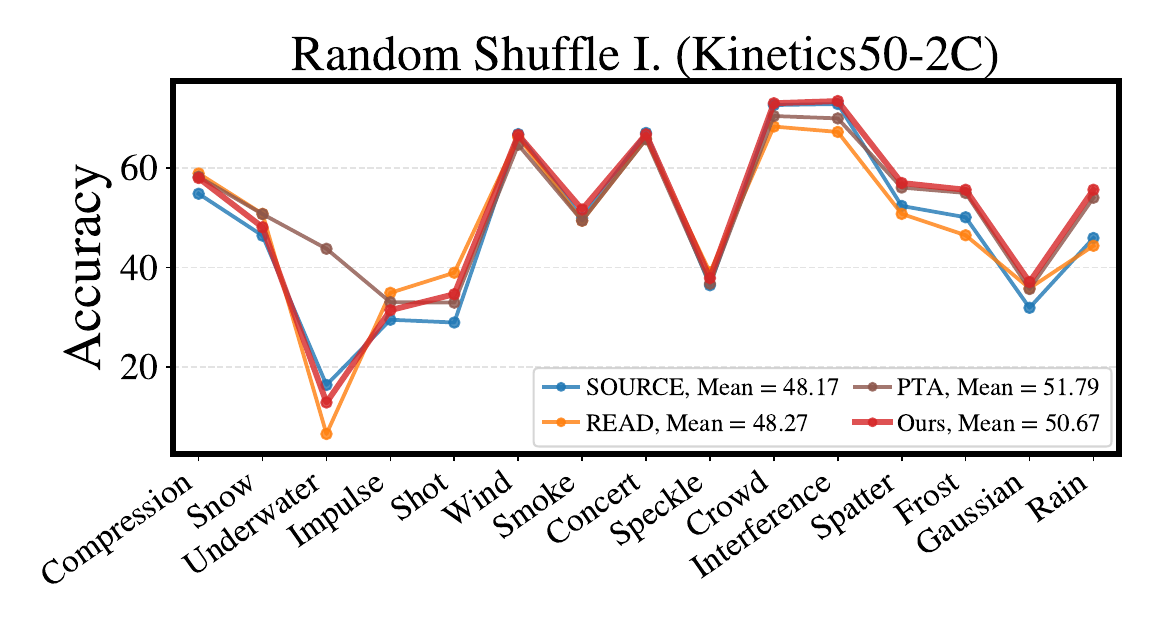}
    \end{subfigure}\hfill
    \begin{subfigure}[t]{0.33\textwidth}
        \centering
        \includegraphics[width=\linewidth,trim={7mm 5mm 5mm 5mm},clip]{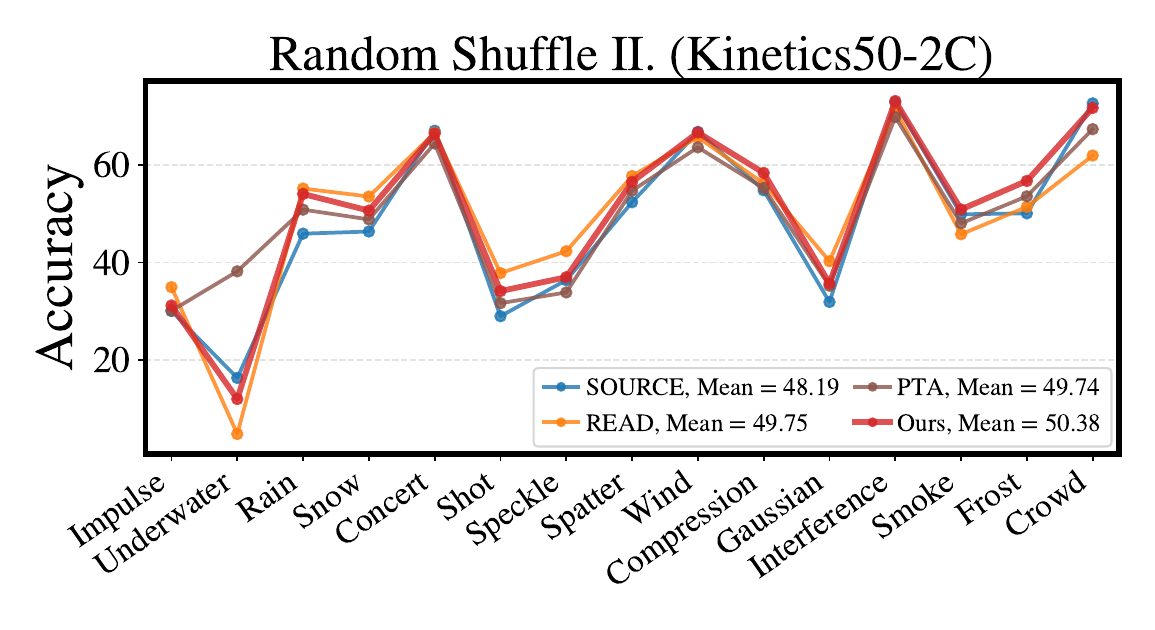}
    \end{subfigure}
    \begin{subfigure}[t]{0.33\textwidth}
        \centering
        \includegraphics[width=\linewidth,trim={7mm 5mm 5mm 5mm},clip]{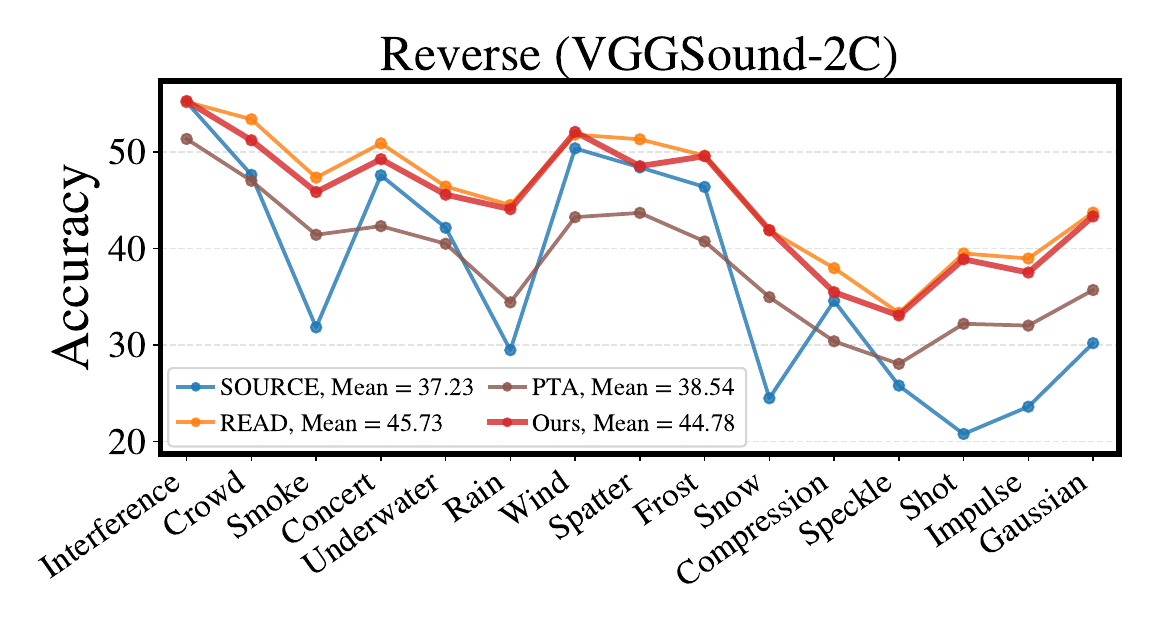}
    \end{subfigure}\hfill
    \begin{subfigure}[t]{0.33\textwidth}
        \centering
        \includegraphics[width=\linewidth,trim={7mm 5mm 5mm 5mm},clip]{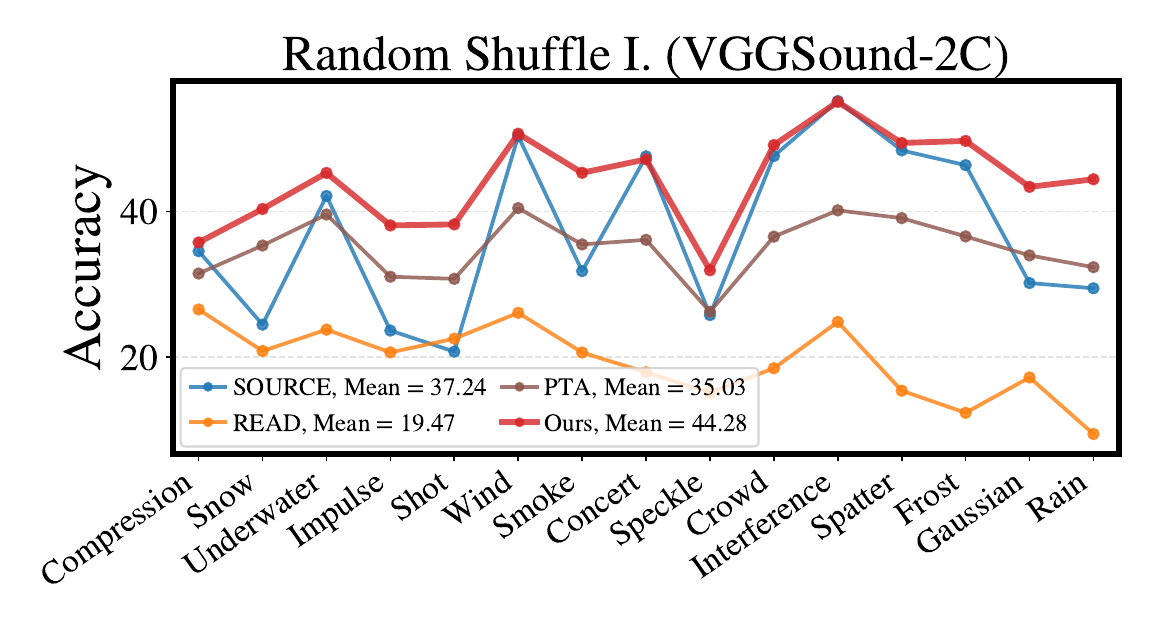}
    \end{subfigure}\hfill
    \begin{subfigure}[t]{0.33\textwidth}
        \centering
        \includegraphics[width=\linewidth,trim={7mm 5mm 5mm 5mm},clip]{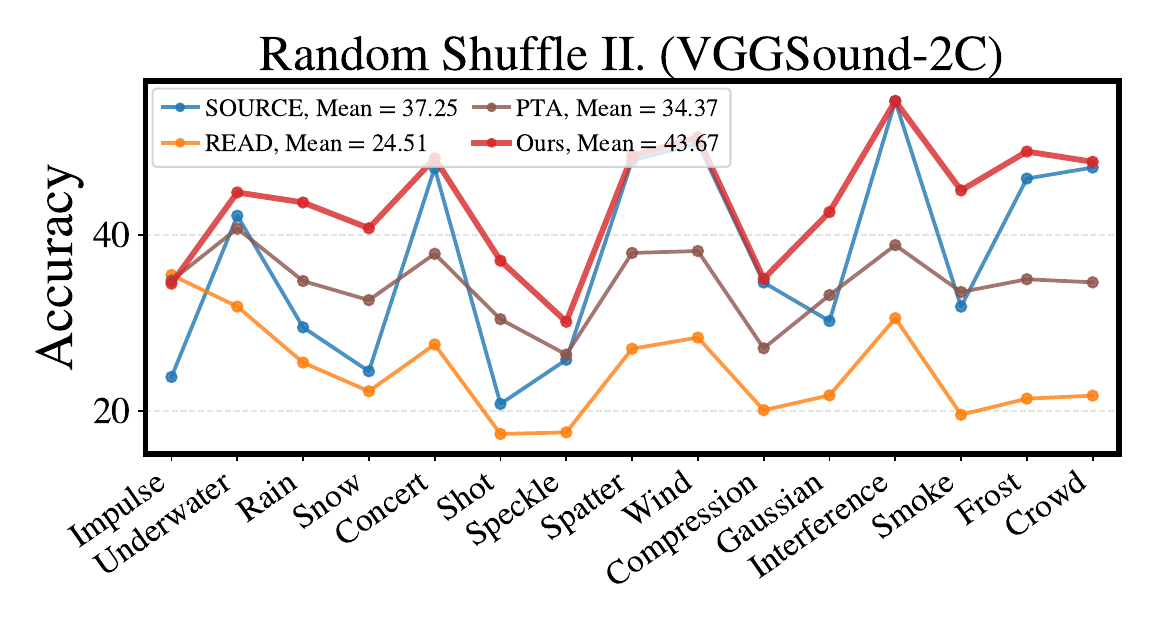}
    \end{subfigure}

    \caption{\textbf{$\framework$ is robust to variations in task ordering.} The top row reports results on Kinetics50-2C, while the bottom row shows results on VGGSound-2C. In each plot, the x-axis denotes the task order in the continual setting, and the y-axis reports task-wise accuracy. The legend summarizes the mean accuracy achieved by each baseline and by our method.}
    \label{fig:task_order}
\end{figure*}

\subsection{Sensitivity to Task Orders}
In real-world continual learning, the sequence of tasks arriving is often unpredictable. To assess the robustness of $\framework$ and competing baselines against this, we evaluate their performance across three alternative task orderings in the challenging bimodal corruption setting. While the standard sequence follows a fixed progression (e.g., $\textit{Gaussian} \rightarrow \textit{Impulse} \rightarrow \dots \rightarrow \textit{Interference}$), we introduce: 1) Reverse: The original sequence is inverted. 2) Random Shuffle I \& II: Two distinct permutations of the 15 bimodal tasks. The objective of this study is to study whether the selective retrieval mechanism in $\framework$ is dependent on a specific curriculum or not. We conduct experiments on Kinetics50-2C and VGGSound-2C and illustrate the results in Figure \ref{fig:task_order}. On average, $\framework$ demonstrates strong robustness to task order variations compared to baselines, particularly in the difficult bimodal setting. Although a performance drop is observed during the initial tasks, continual adaptation allows the model to stabilize over time and converge to comparable accuracy across different task orderings. Overall, the selective retrieval mechanism remains effective without long-term degradation and remains resilient to variations in task ordering.

\section{Limitations and Discussions}
While $\framework$ demonstrates significant advancements in audio-visual CTTA, several avenues remain for further exploration.

\noindent \textbf{Latency and scalibility.} The selective parameter retrieval mechanism described in \S \ref{method:retrieval} performs an $O(N)$ linear search over the buffer $\mathcal{K}$ at each time-step $t$. Although effective, this design introduces non-trivial computational overhead that may become prohibitive as the number of encountered target environments grows. Improving retrieval efficiency or eliminating the need for an explicit buffer remains an open challenge.

\noindent \textbf{Model architectural assumptions.} Our method $\framework$ is developed under the assumption that audio-visual recognition models employ modality-specific encoders followed by a joint encoder for late cross-modal fusion, as commonly adopted in prior architectures \cite{gong2022contrastive, gong2022uavm, georgescu2023audiovisual, huang2023mavil}. Consequently, its applicability to alternative fusion paradigms or more heterogeneous architectures is not guaranteed, limiting its generality.  

\noindent \textbf{Threshold $\tau$.} The method relies on a single hyperparameter $\tau$ to control selective parameter retrieval. In practical test-time deployment settings, hyperparameter tuning is often infeasible or costly \cite{maharana2024texttt}.

\noindent \textbf{Correlation between raw input-level statistics and optimality of the retrieved fusion parameters, and direct validation of retrieval.} Modality-level input statistics $(\mu, \Sigma)$ capture covariate shift, and since cross‑modal attention is conditioned on these inputs, similar shifts induce similar optimal fusion parameters, consistent with prior findings \cite{zhang2024dpcore}.  We store adapted fusion parameters in a buffer and retrieve them using the KL‑divergence similarity. Tracking retrievals over the first 6 tasks of Kinetics50‑C (78 batches per task) shows clear alignment: Shot Noise retrieves from Gaussian 4/78 batches (and the rest are continually adapted), Impulse Noise retrieves from Gaussian (77/78). Defocus Blur uses Gaussian for (1/78) batches, while Glass, Motion, and Zoom Blur always retrieve from Defocus Blur (78/78). This confirms that similar domains retrieve parameters from each other, thereby validating that the parameters previously retrieved by our method correspond to similar domains.

\end{document}